\documentclass{article} 
\usepackage{iclr2027_conference,times}

\usepackage{amsmath,amsfonts,bm}

\def\eqref#1{equation~\ref{#1}}

\def\1{\bm{1}}

\DeclareMathAlphabet{\mathsfit}{\encodingdefault}{\sfdefault}{m}{sl}
\SetMathAlphabet{\mathsfit}{bold}{\encodingdefault}{\sfdefault}{bx}{n}

\usepackage{url}
\usepackage{enumitem}
\usepackage{graphicx}
\usepackage{float}
\usepackage{wrapfig}
\usepackage{afterpage}
\usepackage{capt-of}
\usepackage{booktabs}
\usepackage{tabularx}
\usepackage{multirow}
\usepackage{tcolorbox}
\usepackage{adjustbox}
\usepackage{nicematrix}
\usepackage{tikz}
\usepackage{hyperref}
\definecolor{qualityBetter}{HTML}{E8F1FA}
\definecolor{qualityWorse}{HTML}{FBEFE3}
\definecolor{avmEval}{HTML}{D57577}
\definecolor{avmPref}{HTML}{7085EC}
\definecolor{avmEvalAction}{HTML}{E49B9D}
\definecolor{avmEvalVideo}{HTML}{EDB5B7}
\definecolor{avmPrefAction}{HTML}{8C9EF3}
\definecolor{avmPrefVideo}{HTML}{A3B4F8}
\definecolor{avmAgiTest}{HTML}{E59699}
\definecolor{avmAgiB}{HTML}{EBA7AA}
\definecolor{avmAgiTestVideo}{HTML}{D07478}
\definecolor{avmAgiBVideo}{HTML}{DC888C}
\definecolor{avmRoboT}{HTML}{F0BBBE}
\definecolor{avmRoboD}{HTML}{F3CCCE}
\definecolor{avmGen}{HTML}{F7DCDE}
\newcommand{\avmsource}[2]{\textcolor{#1}{\rule{4pt}{4pt}}\,\ensuremath{#2}}
\newcommand{\avmevalcomposition}{%
  \begin{tikzpicture}[x=1pt,y=1pt]
    \path[use as bounding box] (-35,-35) rectangle (35,35);
    \fill[avmEvalAction,draw=white!85,line width=.4pt] (90.000000:15) arc (90.000000:378.548433:15) -- (378.548433:24.5) arc (378.548433:90.000000:24.5) -- cycle;
    \node[rotate=-35.725784,font=\fontsize{5.8}{6}\selectfont] at (234.274216:19.75) {Action};
    \fill[avmEvalVideo,draw=white!85,line width=.4pt] (378.548433:15) arc (378.548433:450.000000:15) -- (450.000000:24.5) arc (450.000000:378.548433:24.5) -- cycle;
    \node[rotate=-35.725784,font=\fontsize{5.8}{6}\selectfont] at (414.274216:19.75) {Video};
    \fill[avmAgiTest,draw=white!85,line width=.4pt] (90.000000:24.5) arc (90.000000:154.772900:24.5) -- (154.772900:34) arc (154.772900:90.000000:34) -- cycle;
    \fill[avmAgiB,draw=white!85,line width=.4pt] (154.772900:24.5) arc (154.772900:259.109005:24.5) -- (259.109005:34) arc (259.109005:154.772900:34) -- cycle;
    \fill[avmRoboT,draw=white!85,line width=.4pt] (259.109005:24.5) arc (259.109005:320.272104:24.5) -- (320.272104:34) arc (320.272104:259.109005:34) -- cycle;
    \fill[avmRoboD,draw=white!85,line width=.4pt] (320.272104:24.5) arc (320.272104:360.349465:24.5) -- (360.349465:34) arc (360.349465:320.272104:34) -- cycle;
    \fill[avmGen,draw=white!85,line width=.4pt] (360.349465:24.5) arc (360.349465:378.548433:24.5) -- (378.548433:34) arc (378.548433:360.349465:34) -- cycle;
    \fill[avmAgiTestVideo,draw=white!85,line width=.4pt] (378.548433:24.5) arc (378.548433:410.683161:24.5) -- (410.683161:34) arc (410.683161:378.548433:34) -- cycle;
    \fill[avmAgiBVideo,draw=white!85,line width=.4pt] (410.683161:24.5) arc (410.683161:450.000000:24.5) -- (450.000000:34) arc (450.000000:410.683161:34) -- cycle;
    \node[font=\fontsize{5.8}{6.5}\selectfont,inner sep=0pt] at (0,4) {AVM-Eval};
    \node[font=\fontsize{5.8}{6.5}\selectfont,inner sep=0pt] at (0,-4) {67.2k};
  \end{tikzpicture}%
}
\newcommand{\avmprefcomposition}{%
  \begin{tikzpicture}[x=1pt,y=1pt]
    \path[use as bounding box] (-35,-35) rectangle (35,35);
    \fill[avmPrefAction,draw=white!85,line width=.4pt] (90.000000:15) arc (90.000000:201.151910:15) -- (201.151910:24.5) arc (201.151910:90.000000:24.5) -- cycle;
    \node[rotate=55.575955,font=\fontsize{5.8}{6}\selectfont] at (145.575955:19.75) {Action};
    \fill[avmPrefVideo,draw=white!85,line width=.4pt] (201.151910:15) arc (201.151910:450.000000:15) -- (450.000000:24.5) arc (450.000000:201.151910:24.5) -- cycle;
    \node[rotate=55.575955,font=\fontsize{5.8}{6}\selectfont] at (325.575955:19.75) {Video};
    \fill[avmPrefAction,draw=white!85,line width=.4pt] (90.000000:24.5) arc (90.000000:201.151910:24.5) -- (201.151910:34) arc (201.151910:90.000000:34) -- cycle;
    \fill[avmPrefVideo,draw=white!85,line width=.4pt] (201.151910:24.5) arc (201.151910:450.000000:24.5) -- (450.000000:34) arc (450.000000:201.151910:34) -- cycle;
    \node[font=\fontsize{5.8}{6.5}\selectfont,inner sep=0pt] at (0,4) {AVM-Pref};
    \node[font=\fontsize{5.8}{6.5}\selectfont,inner sep=0pt] at (0,-4) {506.3k};
  \end{tikzpicture}%
}
\newcolumntype{Y}{>{\raggedright\arraybackslash}X}
\newsavebox{\oodActionTableBox}
\newsavebox{\agibotRewardTableBox}
\newlength{\referenceTableWidth}
\newlength{\ndtwDeltaWidth}

\newcommand{\us}{\_\allowbreak}

\title{ACWM-RM: A Reference-Free Reward Model\\
from Complementary Mismatch Preferences}

\author{Shibo Hong$^{1}$, Siyu Ren$^{2}$, FengJiao Chen$^{2}$, 
Zhongyuan Peng$^{1}$, Wei Shi$^{2}$, Yixin Cao$^{1}$\\[0.5ex]
\textnormal{$^{1}$College of Computer Science and Artificial Intelligence, Fudan University}
\textnormal{$^{2}$Meituan}
}

\iclrfinalcopy 
\begin{document}

\maketitle

\begin{abstract}
Reinforcement learning has gained attention for improving action-conditioned world models (ACWMs), as it enables reward-driven optimization of visual quality (VQ) and action-following fidelity (AF) beyond flow-matching optimization.
Realizing this potential requires reward signals that reliably rank videos by both VQ and AF.
However, existing rewards capture only one dimension: metric-based rewards assess VQ but require ground-truth videos and do not directly assess AF, while inverse dynamics model (IDM) rewards assess AF without references but do not guarantee VQ.
Moreover, IDMs, typically trained on real videos, may provide unreliable rewards for generated videos containing visual artifacts and physical inconsistencies.
In this work, we introduce the Action and Video Mismatch Benchmark (AVM-Eval), comprising 67.2k pairs that evaluate how reliably IDMs assess AF on real and generated videos.
We then propose ACWM-RM, a reward model that jointly assesses VQ and AF without ground-truth videos during reward computation.
To train it, we first construct the Action and Video Mismatch Preference Dataset (AVM-Pref) with 506.3k pairs providing complementary supervision: action-mismatch pairs supervise AF, while video-mismatch pairs supervise VQ and expose the model to generated videos.
We then pretrain an action-conditioned predictor on real videos to learn forward dynamics before jointly training it with a patch-weighted score head on AVM-Pref to transfer these dynamics to AF assessment on generated videos.
Evaluation on AVM-Eval reveals the limited reliability of IDMs on generated videos, while downstream RL experiments show that ACWM-RM outperforms reward baselines in both criteria.
\end{abstract}

\section{introduction}

Action-conditioned world models (ACWMs)~\citep{yang2023learning,wu2024ivideogpt,zhu2025irasim,liao2025genie} predict videos conditioned on robot actions, supporting embodied policy evaluation and optimization.
For these applications, generated videos must preserve visual quality while faithfully reflecting the commanded actions.
However, standard flow-matching training~\citep{lipman2022flow} does not explicitly optimize these criteria.
This motivates reinforcement learning (RL) such as Flow-GRPO~\citep{liu2025flowgrpo,wu2026roboalign}, which optimizes these criteria using rewards.

\afterpage{
\begin{figure}[!t]
\centering
\setlength{\parskip}{0pt}
\setlength{\abovecaptionskip}{0pt}
\includegraphics[width=\textwidth]{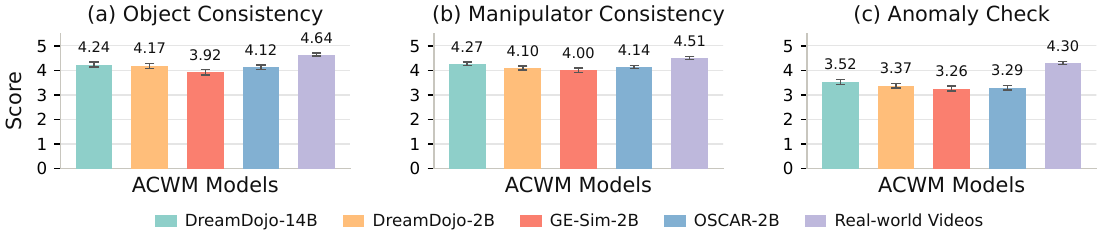}\par
\vspace{-4pt}
\caption{Evaluation scores for generated and real-world videos assigned by GPT-5.6 sol using RBench~\citep{deng2026rethinking} (1: worst; 5: best). Prompts and criteria are detailed in Appendix~\ref{app:prompts_and_criteria}.}
\label{fig:acwm-consistency}
\vspace{-10pt}
\end{figure}
}

The effectiveness of RL, however, depends on reward signals that reliably rank videos generated under the same conditioning input, assigning higher scores to those with better visual quality (VQ) and action-following fidelity (AF). 
Existing reward signals fall into two main categories.
First, full-reference rewards~\citep{wu2026rlvr} use metrics such as PSNR, SSIM, or LPIPS to compare generated videos with the corresponding ground-truth videos, providing VQ supervision but requiring reference videos and not directly assessing AF.
Second, inverse dynamics model (IDM) rewards~\citep{seo2026acid,chen2026robotic} assess AF without requiring reference videos by comparing actions inferred from generated videos with the input actions used to generate those videos, but do not directly assess VQ.
Moreover, IDMs are typically trained on real-world videos, where motion and interactions obey physical laws, but are applied during RL to ACWM-generated videos that may contain visual artifacts and physical inconsistencies, such as robot-arm deformation and interpenetration.
To quantify this issue, we generate 100 videos with each of four ACWMs under identical inputs.
As shown in Figure~\ref{fig:acwm-consistency}, ACWM-generated videos exhibit more visual artifacts and physical inconsistencies than real-world videos.
This gap raises the question of whether strong action-mismatch discrimination on real videos translates to reliable assessment of generated videos.

In this work, we first construct the Action and Video Mismatch Benchmark (AVM-Eval), comprising 67.2k pairs. 
It evaluates how reliably IDMs assess AF on real videos using action-mismatch pairs and on generated videos using video-mismatch pairs.
We then propose a reward model (ACWM-RM) for jointly assessing VQ and AF without ground-truth videos at reward computation time.
To train ACWM-RM, we construct the Action and Video Mismatch Preference Dataset (AVM-Pref) with 506.3k pairs of two types that provide complementary supervision:
Action-mismatch pairs contrast original and perturbed actions for the same real video, providing AF supervision;
Video-mismatch pairs contrast different generated videos under the same actions, providing VQ supervision while exposing the model to the generated-video distribution.
For training strategy, we first pretrain an action-conditioned predictor on features extracted from real robot videos by a frozen V-JEPA 2.1 encoder~\citep{mur2026v}, establishing a dynamics representation for AF evaluation.
We then jointly optimize the predictor and patch-weighted score head on AVM-Pref, transferring the dynamics learned from real-world videos to AF assessment of generated videos.

Our evaluation of four architecturally distinct IDMs on AVM-Eval shows that accurate AF assessment on real videos does not ensure reliable AF evaluation of generated videos.
The best-performing IDM achieves 86.1\% accuracy on action-mismatch pairs from real videos but only 70.0\% on video-mismatch pairs from generated videos.
To evaluate ACWM-RM’s downstream effectiveness as an RL reward, we fine-tune DreamDojo-2B with Flow-GRPO, comparing ACWM-RM against the best-performing IDM on AVM-Eval, a metric-based reward, and a VLM-based reward. 
On held-out tasks, ACWM-RM yields the best VQ and AF, while maintaining VLM-assessed physical consistency (PC) scores close to the original model before RL fine-tuning.

Our main contributions:
(1) We introduce AVM-Eval and show that, across four IDMs, accurate AF assessment on real videos does not ensure reliable AF evaluation of generated videos.
(2) We develop AVM-Pref and ACWM-RM, combining dynamics pretraining with complementary mismatch preferences to assess both criteria without references during reward computation.
(3) We demonstrate that ACWM-RM outperforms reward baselines in VQ and AF when used for RL fine-tuning.

\section{related work}

\subsection{Action-Conditioned World Models}

Action-conditioned world models generate videos conditioned on past observations and actions, simulating the resulting robot interactions to support policy evaluation and planning.
To learn transferable dynamics, DreamDojo~\citep{gao2026dreamdojo} pretrains on human videos using continuous latent actions and adapts to target robots with robot data.
To condition predictions on actions, iVideoGPT~\citep{wu2024ivideogpt} models observations and actions as token sequences with an autoregressive transformer, whereas IRASim~\citep{zhu2025irasim} strengthens action-frame alignment through frame-level conditioning in a diffusion transformer.
These predictive capabilities enable downstream applications: Genie Envisioner~\citep{liao2025genie} uses GE-Sim for action-conditioned simulation and closed-loop policy evaluation.
Such applications require generated videos to preserve both VQ and AF, motivating our study of rewards that jointly assess these properties for ACWM post-training.

\subsection{Reward Modeling for World Models}

Reliable reward signals are essential for effective RL, as they determine which properties of videos are optimized.
For general video generation, VideoReward~\citep{liu2025videoreward} learns human preferences over visual quality and alignment.
For embodied world models based on instructions, ReWorld~\citep{peng2026reworld} and RoboAlign-R1~\citep{wu2026roboalign} assess video quality and instruction alignment without evaluating correspondence with numerical actions.
For action-conditioned world models, RLVR-World~\citep{wu2026rlvr} provides visual rewards through ground-truth video comparisons, while RLIR~\citep{ye2025reinforcement} provides AF rewards by using IDM.
Other dynamics-based rewards target different properties: EVA~\citep{wang2026eva} assesses action executability using IDM.
However, reference-based rewards require ground truth, and IDMs that are accurate on real videos does not ensure reliable evaluation of generated videos or joint assessment of VQ and AF.

\section{Preliminaries}

Given an initial frame $I_0$ and an action chunk $\mathbf{a}=(a_1,\ldots,a_K)$, an ACWM $p_\theta$ predicts video $\hat{\mathbf{I}}=(\hat I_1,\ldots,\hat I_K)$ by sampling $\hat{\mathbf{I}}\sim p_\theta(\cdot\mid c)$, where $c=(I_0,\mathbf{a})$.
To improve VQ and AF for ACWMs, Flow-GRPO is used with reward-based supervision.
The reward $R$ is typically computed using metrics or IDMs.
Metric-based rewards compare generated videos $\hat{\mathbf{I}}$ with ground-truth videos $\mathbf{I}=(I_1,\ldots,I_K)$ using a metric $D$, such as PSNR, SSIM, or MSE.
IDM-based rewards compare actions $\tilde{\mathbf{a}}=(\tilde a_1,\ldots,\tilde a_K)$ inferred from the generated video with the conditioning actions $\mathbf{a}$:
\begin{equation}
R^{\mathrm{metric}}
=s_D\,D\!\left(\hat{\mathbf{I}},\mathbf{I}\right),
\qquad
R^{\mathrm{IDM}}
=-\frac{1}{14K}\sum_{k=1}^{K}\sum_{j=1}^{14}
\ell_{\mathrm{SL1}}\!\left(
\frac{\tilde a_{k,j}-a_{k,j}}{\sigma_j+10^{-8}}
\right).
\end{equation}
Here, $s_D=1$ for higher-is-better metrics and $s_D=-1$ otherwise.
For $R^{\mathrm{IDM}}$, $\ell_{\mathrm{SL1}}$ is the Smooth-L1 loss with threshold 1, and $\sigma_j$ is the standard deviation of action dimension $j$.

To optimize the ACWM with reward $R$, Flow-GRPO first converts the deterministic sampling ODE into an SDE with matching marginals, enabling stochastic sampling in denoising process.
For each $c$, it samples $G$ videos $\{\hat{\mathbf{I}}_i\}_{i=1}^{G}$ from the rollout policy $p_{\theta_{\mathrm{old}}}$ and scores them using $R$.
The resulting scores $R_i$ are normalized into group-relative advantages $\hat A_i=(R_i-\mu_R)/(\sigma_R+\epsilon)$, where $\mu_R$ and $\sigma_R$ are the  group reward mean and standard deviation.
The ACWM is then updated by maximizing:
\begin{equation}
\mathbb{E}\Bigg[\frac{1}{G}\sum_{i=1}^{G}\min\!\Big(r_i(\theta)\hat A_i,\operatorname{clip}\!\big(r_i(\theta),1-\varepsilon,1+\varepsilon\big)\hat A_i\Big)-\beta D_{\mathrm{KL}}\!\left(p_\theta(\cdot\mid c)\,\middle\|\,p_{\mathrm{ref}}(\cdot\mid c)\right)\Bigg],
\end{equation}
where $r_i(\theta)$ is the SDE transition likelihood ratio of $p_\theta$ to $p_{\theta_{\mathrm{old}}}$, $\varepsilon$ is the clipping range, and $\beta$ controls KL regularization toward the frozen reference model $p_{\mathrm{ref}}$.
The clipped objective and KL penalty are averaged over denoising steps; step indices and latent conditioning are omitted for brevity.

\section{Method}
We first describe AVM-Eval construction (Sec.~\ref{sec:avmb}) and then our ACWM-RM, including AVM-Pref construction, model architecture, and two-stage training (Sec.~\ref{sec:reward-model}).

\subsection{Action and Video Mismatch Benchmark}
\label{sec:avmb}

Existing IDMs are normally trained and evaluated on real-world videos.
However, IDM-based rewards are applied to generated videos, whose distributions can differ from those of real-world videos.
Generated videos may contain visual artifacts and physical inconsistencies, including robot-arm deformation and interpenetration, absent from real videos.
To our knowledge, no existing benchmark compares the reliability of IDM-based AF assessment on real and generated videos.

To address this gap, we construct AVM-Eval using a pairwise evaluation protocol~\citep{lambert2025rewardbench}.
AVM-Eval compares the reliability of AF assessment on real and ACWM-generated videos, using \textbf{action-mismatch pairs} and \textbf{video-mismatch pairs}, respectively.
We next describe the source data preparation and benchmark construction.

\subsubsection{Source data preparation}
\label{sec:source-data-preparation}

We use AgiBotWorld~\citep{bu2025agibot}, a large-scale bimanual robot dataset with real-world videos, as one source dataset, and partition its data by task into three subsets: \(\mathrm{Agi_A}\), \(\mathrm{Agi_B}\), and \(\mathrm{Agi_C}\).
Covering 36 tasks, \(\mathrm{Agi_A}\) is split at the episode level into 34.0k training episodes (\(\mathrm{Agi_A^{train}}\)) for IDM, ACWM-RM predictor pretraining, and AVM-Pref construction, and 2.2k testing episodes (\(\mathrm{Agi_A^{test}}\)) for constructing seen-task evaluation pairs in AVM-Eval.
Using \(\mathrm{Agi\_B}\), which contains 10.0k episodes across 30 tasks, we construct pairs for both unseen-task evaluation in AVM-Eval and RL fine-tuning.
To evaluate RL-optimized ACWMs, we use \(\mathrm{Agi_C}\), which contains 1.5k episodes across 10 tasks.
Additionally, to evaluate the embodiment and scene generalization of the IDMs in AVM-Eval, we use other bimanual robot datasets from Genie-Sim (\(\mathrm{Gen}\))~\citep{yin2026geniesim30}, RoboDojo (\(\mathrm{Robo_D}\))~\citep{chen2026robodojounifiedsimandrealbenchmark}, and RoboTwin (\(\mathrm{Robo_T}\))~\citep{mu2025robotwindualarmrobotbenchmark}.
Table~\ref{tab:benchmark_composition} shows statistics of AVM-Eval and AVM-Pref.
Appendix~\ref{app:statistics} provides more statistics.
Across all datasets, we subsample each episode to 8 fps and randomly extract two 13-frame video clips for all experiments, each with corresponding end-effector states $\mathbf{s}=(s_0,\ldots,s_{12})$ and action chunks $\mathbf{a}=(a_1,\ldots,a_{12})$.

\begin{table*}[t]
\centering\small
\setlength{\belowcaptionskip}{3pt}
\caption{Statistics of AVM-Eval and AVM-Pref. Abbreviations:
N-P/S-P, numerical/semantic perturbations; Intra/Cross, intra-model and cross-model pairs.
Video source: Rel., real-world; Rec., simulator-recorded; Gen., generated.
Criteria: AF, action-following fidelity; VQ, visual quality.}
\label{tab:benchmark_composition}
\setlength{\tabcolsep}{2pt}
\renewcommand{\arraystretch}{1.0}
\setlength{\aboverulesep}{1.5pt}
\setlength{\belowrulesep}{2pt}
\begin{minipage}[c]{0.199\textwidth}
\centering
\setlength{\parskip}{0pt}
\scalebox{1.10}{\avmevalcomposition}\par\vspace{-2pt}
\scalebox{1.10}{\avmprefcomposition}
\end{minipage}\hspace{2pt}%
\scalebox{0.98}{%
\begin{minipage}[c]{0.81\textwidth}
\centering\footnotesize
\begin{adjustbox}{max width=\linewidth}
\begin{NiceTabular*}{\linewidth}{@{\extracolsep{\fill}}c|lcccccccc@{}}
\toprule
 & \multirow{2}{*}{Source} & \multirow{2}{*}{Tasks} & \multirow{2}{*}{Episodes} & \multicolumn{2}{c}{Action Pairs} & \multicolumn{2}{c}{Video Pairs} & \multirow{2}{*}{Video} & \multirow{2}{*}{Criteria} \\
\cmidrule(lr){5-6}\cmidrule(lr){7-8}
 & & & & N-P & S-P & Intra & Cross & & \\
\midrule
\multirow{7}{*}{\rotatebox{90}{AVM-Eval}} & \avmsource{avmAgiTest}{\mathrm{Agi_A^{test}}} & 36 & 2.2k & 4.9k & 7.2k & -- & -- & Rel. & AF \\
 & \avmsource{avmAgiB}{\mathrm{Agi_B}} & 30 & 10.0k & 12.5k & 7.0k & -- & -- & Rel. & AF \\
 & \avmsource{avmRoboT}{\mathrm{Robo_T}} & 50 & 2.2k & 4.4k & 7.0k & -- & -- & Rec. & AF \\
 & \avmsource{avmRoboD}{\mathrm{Robo_D}} & 35 & 1.5k & 2.6k & 4.9k & -- & -- & Rel. & AF \\
 & \avmsource{avmGen}{\mathrm{Gen}} & 19 & 2.3k & 1.2k & 2.2k & -- & -- & Rec. & AF \\
\cmidrule(l){2-10}
 & \avmsource{avmAgiTestVideo}{\mathrm{Agi_A^{test}}} & 36 & 2.2k & -- & -- & 3.0k & 3.0k & Gen. & AF \\
 & \avmsource{avmAgiBVideo}{\mathrm{Agi_B}} & 30 & 10.0k & -- & -- & 4.3k & 3.0k & Gen. & AF \\
\midrule
\multirow{2}{*}[1.2pt]{\raisebox{-.5\height}[0pt][0pt]{\rotatebox{90}{\fontsize{7}{8}\selectfont\shortstack{AVM\\Pref}}}} & \rule{0pt}{11pt}\avmsource{avmPrefAction}{\mathrm{Agi_A^{train}}} & 36 & 34.0k & 63.5k & 92.8k & -- & -- & Rel. & AF \\
 & \rule{0pt}{11pt}\avmsource{avmPrefVideo}{\mathrm{Agi_A^{train}}} & 36 & 34.0k & -- & -- & 250.0k & 100.0k & Gen. & VQ \\
\bottomrule
\end{NiceTabular*}
\end{adjustbox}
\end{minipage}%
}
\vspace{-25pt}
\end{table*}

\subsubsection{AVM-Eval Construction}
\label{sec:preference-pair-construction}
AVM-Eval comprises two complementary types of pairs:
A video-mismatch pair holds the action chunk fixed and contrasts two videos, whereas an action-mismatch pair holds the video fixed and contrasts two action chunks. 
Together, these two types constitute 67.2k pairs in AVM-Eval.

\textbf{Action-mismatch pairs (53.9k).}
To evaluate IDM-based AF assessment on real videos, we construct action-mismatch pairs.
Each pair is represented as $(I_0,\mathbf{a}^{+},\mathbf{a}^{-},\mathbf{I})$, where $\mathbf{a}^{+}$ is the action chunk recorded with $\mathbf{I}$ and $\mathbf{a}^{-}$ is a perturbed counterpart.
We generate $\mathbf{a}^{-}$ using two types of perturbations: Numerical Perturbations (N-P) and Semantic Perturbations (S-P).
N-P probe sensitivity to low-level numerical errors in end-effector translation and rotation magnitudes or directions, gripper opening degree, and action ordering.
S-P probe sensitivity to high-level semantic errors through position offsets during grasping or placement, incorrect gripper timing, and mid-motion drift or freeze.
We also perturb action chunks from $\mathrm{Gen}$, $\mathrm{Robo_D}$, and $\mathrm{Robo_T}$ to assess IDM generalization across embodiments and scenes.
Detailed perturbation procedures are provided in Appendix~\ref{app:action_perturbation}.

\textbf{Video-mismatch pairs (13.3k).}
To evaluate AF on generated videos, we considered constructing action-mismatch pairs in the same way.
However, ACWMs cannot guarantee that generated videos faithfully follow the action chunk, so a perturbed action chunk may be more consistent with the generated video than the original action chunk, reversing the intended preference label.
We therefore construct video-mismatch pairs by fixing the action chunk and ranking generated videos using WorldArena's nDTW~\citep{shang2026worldarena} for AF, with additional filtering to limit the influence of visual quality (Appendix~\ref{app:ndtw-pair-filtering}).
We use four ACWMs with varying AF performance: DreamDojo-2B (DD-2B), DreamDojo-14B (DD-14B)~\citep{gao2026dreamdojo}, GE-Sim-2B (GE-2B), and OSCAR-2B (OS-2B)~\citep{wu2026oscar}.
Under the same $I_0$ and $\mathbf{a}$, each ACWM independently samples six 12-frame videos using the stochastic SDE sampler adopted by Flow-GRPO.
Within each ACWM, we pair a higher-nDTW video \(\hat{\mathbf{I}}^{+}\) with a lower-nDTW video \(\hat{\mathbf{I}}^{-}\) to form an \textbf{intra-model pair} \((I_0,\mathbf{a},\hat{\mathbf{I}}^{+},\hat{\mathbf{I}}^{-})\).
We rank videos generated by any two ACWMs under the same input to construct \textbf{cross-model pairs}.

\subsection{Reward Model for Action-Conditioned World Models}
\label{sec:reward-model}
\begin{figure}[t]
\centering
\setlength{\abovecaptionskip}{4pt}
\includegraphics[width=0.95\textwidth,pagebox=cropbox,clip]{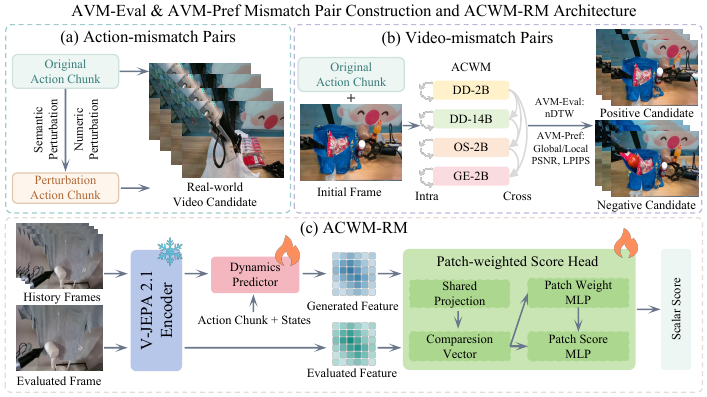}\par
\caption{Overview of (a,b) action-mismatch and video-mismatch pair construction for AVM-Eval and AVM-Pref, and (c) ACWM-RM reward computation without reference videos.}
\label{fig:main}
\vspace{-6pt}
\end{figure}
Existing rewards provide only partial supervision, with metric-based rewards also requiring ground-truth videos.
We therefore propose ACWM-RM, a reference-free reward model that jointly assesses VQ and AF.
We next describe the construction of its training dataset, termed AVM-Pref (Sec.~\ref{action-video-preference-dataset}), followed by the architecture and the two-stage training strategy (Sec.~\ref{reward-model-architectural}).

\subsubsection{AVM-Pref Construction}
\label{action-video-preference-dataset}

To train ACWM-RM with the Bradley-Terry loss, we construct the Action and Video Mismatch Preference Dataset (AVM-Pref) from $\mathrm{Agi_A^{train}}$, comprising 506.3k action-mismatch and video-mismatch pairs that provide complementary supervision.
\textbf{Action-mismatch pairs (156.3k)}:
We use the same numerical and semantic action perturbations as AVM-Eval to provide AF supervision.
\textbf{Video-mismatch pairs (350k)}:
We follow AVM-Eval to generate videos from four ACWMs and construct intra-model and cross-model pairs, but use four visual quality metrics (global and local PSNR and LPIPS) instead of nDTW to provide VQ supervision.
Since action-mismatch pairs already provide broader AF supervision than nDTW’s 2D trajectory alignment, including translation, rotation, and gripper opening, we use video-mismatch pairs for complementary VQ supervision.
Although action-mismatch pairs use real videos and thus differ from the generated-video distribution, our empirical results show that combining both types improves AF assessment on generated videos.
Finally, to reduce sampling bias, we aim for comparable pair counts across models and tasks for intra-model pairs and across model combinations and tasks for cross-model pairs.
Appendix~\ref{app:video-mismatch-pairs-for-AVM-Pre} details how the four metrics are used, while Appendix~\ref{app:statistics} provides AVM-Pref statistics.

\subsubsection{Reward Model Architecture and Training Strategy}
\label{reward-model-architectural}

ACWM-RM scores a candidate video conditioned on an action chunk and its associated state sequence, producing a scalar reward for VQ and AF.
It consists of a frozen V-JEPA 2.1 encoder $E$, a bimanual action-conditioned predictor $P_{\phi}$, and a patch-weighted score head $h_{\psi}$.
Training proceeds in two stages: dynamics pretraining optimizes $P_{\phi}$ on real-world videos, while preference training jointly optimizes $P_{\phi}$ and $h_{\psi}$ on AVM-Pref to adapt the learned dynamics for AF assessment.

\textbf{Dynamics pretraining.}
Following V-JEPA 2-AC~\citep{assran2025v}, we pretrain $P_{\phi}$ with $E$ frozen.
Unlike the original single-arm setting, our predictor takes bimanual actions.
For bimanual inputs, we express actions and states in a coordinate system aligned with the head-camera axes, normalize them, project them to the predictor's hidden dimension, and add a learnable identity embedding for each arm.
At each time step $t$, tokens are ordered as $\big[u^s_{t,L},u^a_{t,L},u^s_{t,R},u^a_{t,R},U^z_t\big]$, where $u^s_{t,r}$ and $u^a_{t,r}$ denote state and action tokens for arm $r\in\{L,R\}$, and $U^z_t$ contains projected frame features from $E$.
Appendix~\ref{app:predictor} provides further details and loss functions.

\textbf{Preference training.}
We jointly train $P_{\phi}$ and $h_{\psi}$ to rank pairs by comparing predicted and encoded frame features and aggregating the resulting patch scores.
For each candidate video $\mathbf{I}'$ from either $\mathbf{I}$ or $\hat{\mathbf{I}}$, $P_{\phi}$ predicts a layer-normalized representation $z_t^P$ at time step $t$ from the candidate's encoded frames $I'_{0:t-1}$, actions $a'_{1:t}$, and states $s'_{0:t-1}$, while $E$ encodes the evaluated frame as $z_t^E=\operatorname{LN}(E(I_t'))$.
Here, $s'_{0:t-1}$ is reconstructed from the candidate's actions and initial state $s_0$.
To avoid accumulating rollout errors, the predictor conditions on the candidate's own history rather than recursively predicted features.
For scoring, $h_{\psi}$ compares $z_t^P$ and $z_t^E$ patch by patch, using the spatial correspondence between their $N$ patch features.
Specifically, for each patch $n$, a shared projection maps both features to a common space, yielding $\bar z^P_{t,n}=W_{\mathrm{proj}}z^P_{t,n}$ and $\bar z^E_{t,n}=W_{\mathrm{proj}}z^E_{t,n}$.
The projected features are then combined into a comparison vector:
\begin{equation}
    c_{t,n}=\big[\bar z^P_{t,n}-\bar z^E_{t,n};\,
    |\bar z^P_{t,n}-\bar z^E_{t,n}|;\,
    (\bar z^P_{t,n}-\bar z^E_{t,n})^2;\,
    \bar z^P_{t,n}\odot\bar z^E_{t,n};\,
    \cos(\bar z^P_{t,n},\bar z^E_{t,n})\big],
    \label{eq:acwm-comparison-feature}
\end{equation}
which combines feature differences, elementwise product, and cosine similarity.

Different regions contribute unequally to VQ and AF assessment.
Thus, we use two MLPs to predict patch scores and spatial weight logits, enabling adaptive aggregation of evidence for both criteria:
\begin{equation}
    \ell_{t,n}=f_{\mathrm{score}}(c_{t,n}),\qquad
    \tilde\alpha_{t,n}=f_{\mathrm{weight}}(c_{t,n}).
    \label{eq:acwm-score-and-weight-mlps}
\end{equation}
We apply softmax over patches to obtain weights, compute each frame's score as a weighted sum of patch scores, and average frame-level scores over $K$ future frames to obtain the candidate reward:
\begin{equation}
     R_{\phi,\psi}({\mathbf{I'}},\mathbf{a'},{\mathbf{s'}})
    =\frac{1}{K}\sum_{t=1}^{K}\sum_{n=1}^{N}\alpha_{t,n}\ell_{t,n},
    \qquad
    \alpha_{t,n}=\frac{\exp(\tilde\alpha_{t,n})}{\sum_{n'=1}^{N}\exp(\tilde\alpha_{t,n'})}.
    \label{eq:acwm-clip-score}
\end{equation}
We score both candidates with shared parameters and minimize the Bradley-Terry loss to favor the preferred candidate, keeping $E$ frozen.
Appendix~\ref{app:preference-training} provides the loss and training details.

\section{experiments}
\label{sec:experiments}
We first describe experimental setup (Sec.~\ref{sec:experimental_settings}) and evaluate IDMs and ACWM-RM on AVM-Eval (Sec.~\ref{sec:reward_model_performance_on_AVM-Eval}).
We then compare Flow-GRPO reward signals (Sec.~\ref{sec:reward_model_performance_for_Flow-GRPO}) and conduct ablations (Sec.~\ref{sec:ablation_study}).

\subsection{Experimental Setup}
\label{sec:experimental_settings}

\textbf{Baselines.}
On AVM-Eval, we compare ACWM-RM with four architecturally distinct IDMs:
EVA-IDM ($\mathrm{EVA_{I}}$)~\citep{wang2026eva}, which uses a convolutional backbone;
GR00T-IDM ($\mathrm{GRT_{I}}$)~\citep{bjorck2025gr00t}, which uses a diffusion-based model;
TAP-IDM ($\mathrm{TAP_{I}}$)~\citep{shi2026taskagnostic}, a vision-language model equipped with an action expert;
and J-IDM ($\mathrm{J_{I}}$)~\citep{li2026vera}, which recovers actions by inverting a predicted per-pixel Jacobian field.
This evaluates how reliably IDMs with different architectures transfer from real-video training to generated-video evaluation.

For Flow-GRPO training of ACWMs, we compare ACWM-RM with IDM-based, metric-based, and VLM-based rewards.
The IDM-based baseline uses the best-performing IDM on AVM-Eval.
The metric-based baseline combines the four visual quality metrics used in AVM-Pref following the fusion procedure of RLVR-World~\citep{wu2026rlvr}.
We further explore whether VLM-as-a-Judge~\citep{hong2025frabench} can provide effective rewards for ACWM optimization.
Specifically, we use Qwen3-VL-30B-A3B-Instruct to score videos using embodiment-specific prompts and criteria from RBench.
Prompts and criteria definitions are provided in the Appendix~\ref{app:prompts_and_criteria}.

\textbf{Training setup.}
We train the four IDMs on $\mathrm{Agi_A^{train}}$ using their official pretrained backbones and implementations, following the original training paradigm of predicting actions from real-world videos.
Each IDM and ACWM-RM has a nominal budget of approximately 7.12M 13-frame video draws, including both ACWM-RM training stages.
For Flow-GRPO, we fine-tune DreamDojo-2B on $\mathrm{Agi_B}$ and evaluate it on $\mathrm{Agi_C}$.
Implementation details for the IDMs, ACWM-RM, and Flow-GRPO are provided in Appendices~\ref{app:idm-training-setups}, \ref{app:reward-model-implementation}, and~\ref{app:flow_grpo_details}, respectively.

\textbf{Evaluation protocol.}
We evaluate IDMs using pairwise accuracy on AVM-Eval.
For each ACWM optimized with Flow-GRPO, we repeat video generation five times and report the mean and standard deviation of the following metrics across runs:
(1) VQ, measured by PSNR, LPIPS, SSIM, MSE, and FVD;
(2) AF, measured by nDTW, a proxy based on the alignment of generated and ground-truth 2D manipulator trajectories;
and (3) physical consistency, assessed by GPT-5.6 sol using the same prompt and criteria as the VLM-based reward, covering Object Consistency (OC), Manipulator Consistency (MC), and Anomaly Check (AC).

\begin{table*}[!t]
\centering
\caption{Pairwise accuracy on AVM-Eval within the AgiBot domain. $\mathrm{Agi_A^{test}}$ evaluates seen tasks and $\mathrm{Agi_B}$ evaluates task generalization. N-P and S-P denote numerical and semantic perturbations, respectively.  Intra and Cross denote Intra-model and Cross-model pairs, respectively.}
\label{tab:agibot-reward-accuracy}
\begin{lrbox}{\agibotRewardTableBox}%
\begin{adjustbox}{max width=\textwidth}
\begin{NiceTabular}{lcc!{\vrule}cc!{\vrule}c|cc!{\vrule}cc!{\vrule}c}
\toprule
\Block[m]{3-1}{Model} & \multicolumn{5}{c|}{Action-mismatch Pair} & \multicolumn{5}{c}{Video-mismatch Pair} \\
\cmidrule(lr){2-6}\cmidrule(lr){7-11}
 & \multicolumn{2}{c!{\vrule}}{$\mathrm{Agi_A^{test}}$} & \multicolumn{2}{c!{\vrule}}{$\mathrm{Agi_B}$} & \Block[m]{2-1}{Overall} & \multicolumn{2}{c!{\vrule}}{$\mathrm{Agi_A^{test}}$} & \multicolumn{2}{c!{\vrule}}{$\mathrm{Agi_B}$} & \Block[m]{2-1}{Overall} \\
\cmidrule{2-5}\cmidrule{7-10}
 & N-P & S-P & N-P & S-P & & Intra & Cross & Intra & Cross & \\
\midrule
$\mathrm{EVA_{I}}$ & \underline{92.1} & \underline{81.6} & \underline{90.6} & \underline{80.2} & \underline{86.1} & \underline{69.3} & \underline{73.8} & \underline{70.2} & \underline{66.8} & \underline{70.0} \\
$\mathrm{GRT_{I}}$ & 74.7 & 70.9 & 75.1 & 61.1 & 70.5 & 52.4 & 52.7 & 51.9 & 52.3 & 52.3 \\
$\mathrm{TAP_{I}}$ & 79.1 & 67.3 & 76.2 & 63.6 & 71.6 & 58.5 & 58.0 & 55.7 & 58.3 & 57.6 \\
$\mathrm{J_{I}}$ & 76.7 & 62.2 & 77.0 & 63.7 & 69.9 & 58.3 & 57.9 & 57.8 & 55.7 & 57.4 \\
\midrule
Ours & \textbf{97.9} & \textbf{86.1} & \textbf{97.7} & \textbf{84.7} & \textbf{91.6} & \textbf{90.5} & \textbf{93.3} & \textbf{86.2} & \textbf{84.3} & \textbf{88.6} \\
\bottomrule
\end{NiceTabular}
\end{adjustbox}%
\end{lrbox}%
\global\referenceTableWidth=\wd\agibotRewardTableBox
\usebox{\agibotRewardTableBox}
\vspace{-10pt}
\end{table*}

\begin{table*}[!t]
\centering
\begin{minipage}[t]{0.60\referenceTableWidth}
\centering\small
\caption{Pairwise accuracy on action-mismatch pairs, testing embodiment and scene generalization. N-P and S-P denote numerical and semantic perturbations.}
\label{tab:ood-action-mismatch-accuracy}
\end{minipage}\hspace{0.025\referenceTableWidth}%
\begin{minipage}[t]{0.375\referenceTableWidth}
\centering\small
\caption{Action-following fidelity on $\mathrm{Agi_C}$ (nDTW $\times 100$). Colors follow Table~\ref{tab:flow-grpo-quality-consistency}.}
\label{tab:flow-grpo-action-consistency}
\end{minipage}
\par\nointerlineskip
\begin{lrbox}{\oodActionTableBox}%
\small%
\setlength{\tabcolsep}{3.5pt}%
\begin{adjustbox}{width=0.60\referenceTableWidth}
\begin{NiceTabular}{lcc!{\vrule}cc!{\vrule}cc!{\vrule}c}
\toprule
\Block[m]{2-1}{Model} & \multicolumn{2}{c!{\vrule}}{$\mathrm{Gen}$} & \multicolumn{2}{c!{\vrule}}{$\mathrm{Robo_T}$} & \multicolumn{2}{c!{\vrule}}{$\mathrm{Robo_D}$} & \Block[m]{2-1}{Overall} \\
\cmidrule{2-7}
 & N-P & S-P & N-P & S-P & N-P & S-P & \\
\midrule
$\mathrm{EVA_{I}}$ & \underline{61.5} & 50.9 & 69.5 & \underline{70.6} & 56.5 & 59.5 & 61.4 \\
$\mathrm{GRT_{I}}$ & 50.9 & 49.7 & 61.6 & 63.5 & 59.4 & 59.5 & 57.4 \\
$\mathrm{TAP_{I}}$ & 58.9 & 52.4 & \underline{74.6} & 54.2 & \textbf{66.7} & \textbf{62.9} & \underline{61.6} \\
$\mathrm{J_{I}}$ & 59.9 & \underline{54.2} & 63.5 & 61.9 & 58.1 & 55.5 & 58.9 \\
\midrule
Ours & \textbf{65.9} & \textbf{55.5} & \textbf{77.8} & \textbf{72.4} & \underline{66.2} & \underline{60.9} & \textbf{66.5} \\
\bottomrule
\end{NiceTabular}
\end{adjustbox}%
\end{lrbox}
\begin{minipage}[t]{0.60\referenceTableWidth}
\vspace{0pt}
\centering
\usebox{\oodActionTableBox}
\end{minipage}\hspace{0.025\referenceTableWidth}%
\begin{minipage}[t]{0.375\referenceTableWidth}
\vspace{0pt}
\centering\small
\setlength{\tabcolsep}{0pt}
\renewcommand{\arraystretch}{1.12}
\settowidth{\ndtwDeltaWidth}{{\scriptsize\ensuremath{\pm 0.00}}}
\newcommand{\ndtwcell}[3]{#1\hspace{0.12em}{\scriptsize\textcolor{#2}{\ensuremath{#3}}}}
\begin{adjustbox}{width=\linewidth,totalheight=\dimexpr\ht\oodActionTableBox+\dp\oodActionTableBox\relax}
\begin{NiceTabular*}{\linewidth}{@{}w{l}{0.44\linewidth}w{c}{0.56\linewidth}@{}}
\CodeBefore
  \cellcolor{qualityBetter}{7-2}
  \cellcolor{qualityWorse}{4-2,5-2,6-2}
\Body
\toprule
\Block[m]{2-1}{Method} & \Block[m]{2-1}{nDTW $\uparrow$} \\
 & \\
\midrule
Origin & \ndtwcell{14.84}{black!65}{\pm 0.24} \\
IDM-based & \ndtwcell{13.21}{black!65}{\pm 0.38} \\
VLM-based & \ndtwcell{14.00}{black!65}{\pm 0.41} \\
Metric-based & \ndtwcell{14.05}{black!65}{\pm 0.33} \\
\midrule
\textbf{Ours} & \ndtwcell{\textbf{16.41}}{black!65}{\pm 0.22} \\
\bottomrule
\end{NiceTabular*}
\end{adjustbox}
\end{minipage}
\vspace{-8pt}
\end{table*}

\subsection{IDM Performance on AVM-Eval}
\label{sec:reward_model_performance_on_AVM-Eval}

Table~\ref{tab:agibot-reward-accuracy} examines whether IDMs trained on real videos transfer to AF assessment on generated videos.
On action-mismatch pairs from real videos, IDM accuracies range from 69.9 to 86.1, with $\mathrm{EVA_{I}}$ performing best.
On video-mismatch pairs from ACWM-generated videos, however, $\mathrm{EVA_{I}}$ reaches only 70.0, while the other three IDMs score 52.3 to 57.6.
These results show that strong action-mismatch performance on real videos does not ensure reliable AF evaluation on generated videos.
Therefore, before using an IDM as a reward for ACWM training, its ability to accurately infer actions from generated videos should be evaluated.
Table~\ref{tab:ood-action-mismatch-accuracy} shows limited IDM generalization across embodiments and scenes: overall action-mismatch accuracy ranges from 57.4 to 61.6.

\begin{table*}[!t]
\centering\small
\caption{Visual quality and physical consistency of DreamDojo on $\mathrm{Agi_C}$ after Flow-GRPO with different reward signals, compared with the
  pre-RL model (Origin). LPIPS, SSIM, and MSE are reported $\times 100$, and physical consistency scores range from 1 to 5.
  \textcolor{qualityBetter}{\rule{6pt}{6pt}}\,Better\quad
  \textcolor{qualityWorse}{\rule{6pt}{6pt}}\,Worse than Origin.}
\label{tab:flow-grpo-quality-consistency}
\label{tab:flow-grpo-visual-quality}
\label{tab:flow-grpo-physical-consistency}
\setlength{\tabcolsep}{1.5pt}
\renewcommand{\arraystretch}{1.05}
\newcommand{\qualitycell}[2]{#1\hspace{0.10em}{\scriptsize\textcolor{black!65}{\ensuremath{\pm\,}\makebox[2.0em][l]{#2}}}}
\begin{adjustbox}{width=\textwidth}
\begin{NiceTabular}{@{}lccccc@{\hspace{7pt}}cccc@{}}
\CodeBefore
  \cellcolor{qualityBetter}{5-2,5-6,6-2,6-3,6-4,6-5,6-6,7-2,7-3,7-4,7-5,7-6}
  \cellcolor{qualityWorse}{4-2,4-3,4-4,4-5,4-6,4-7,4-8,4-9,4-10,5-3,5-4,5-7,5-8,5-9,5-10,6-7,6-8,6-9,6-10,7-7,7-8,7-9,7-10}
\Body
\toprule
\Block[m]{2-1}{Method} & \multicolumn{5}{c}{Visual Quality} & \multicolumn{4}{c}{Physical Consistency} \\
\cmidrule(lr){2-6}\cmidrule(lr){7-10}
 & PSNR $\uparrow$ & LPIPS $\downarrow$ & SSIM $\uparrow$ & MSE $\downarrow$ & FVD $\downarrow$ & OC $\uparrow$ & MC $\uparrow$ & AC $\uparrow$ & Overall $\uparrow$ \\
\midrule
Origin & \qualitycell{23.13}{0.06} & \qualitycell{3.66}{0.04} & \qualitycell{88.67}{0.02} & \qualitycell{0.49}{0.0} & \qualitycell{18.31}{1.09} & \qualitycell{\textbf{4.25}}{0.10} & \qualitycell{\textbf{4.31}}{0.29} & \qualitycell{\textbf{3.43}}{0.06} & \qualitycell{\textbf{4.00}}{0.10} \\
IDM-based & \qualitycell{22.82}{0.08} & \qualitycell{4.27}{0.04} & \qualitycell{88.04}{0.11} & \qualitycell{0.58}{0.0} & \qualitycell{31.31}{0.07} & \qualitycell{4.19}{0.17} & \qualitycell{4.16}{0.37} & \qualitycell{3.42}{0.09} & \qualitycell{3.92}{0.14} \\
VLM-based & \qualitycell{24.03}{0.17} & \qualitycell{3.75}{0.12} & \qualitycell{88.41}{0.28} & \qualitycell{0.49}{0.0} & \qualitycell{18.20}{1.08} & \qualitycell{4.12}{0.17} & \qualitycell{4.07}{0.32} & \qualitycell{3.38}{0.06} & \qualitycell{3.86}{0.12} \\
Metric-based & \qualitycell{24.20}{0.09} & \qualitycell{3.63}{0.06} & \qualitycell{89.62}{0.19} & \qualitycell{0.48}{0.0} & \qualitycell{17.56}{0.63} & \qualitycell{4.11}{0.22} & \qualitycell{4.07}{0.36} & \qualitycell{3.34}{0.12} & \qualitycell{3.84}{0.15} \\
\midrule
\textbf{Ours} & \qualitycell{\textbf{24.34}}{0.06} & \qualitycell{\textbf{3.51}}{0.05} & \qualitycell{\textbf{89.86}}{0.13} & \qualitycell{\textbf{0.46}}{0.0} & \qualitycell{\textbf{17.10}}{0.56} & \qualitycell{4.20}{0.12} & \qualitycell{4.17}{0.22} & \qualitycell{3.42}{0.05} & \qualitycell{3.93}{0.08} \\
\bottomrule
\end{NiceTabular}
\end{adjustbox}
\par\vspace{4pt}
\begin{minipage}{\textwidth}
\centering
\setlength{\abovecaptionskip}{4pt}
\includegraphics[width=0.96\textwidth,pagebox=cropbox,clip]{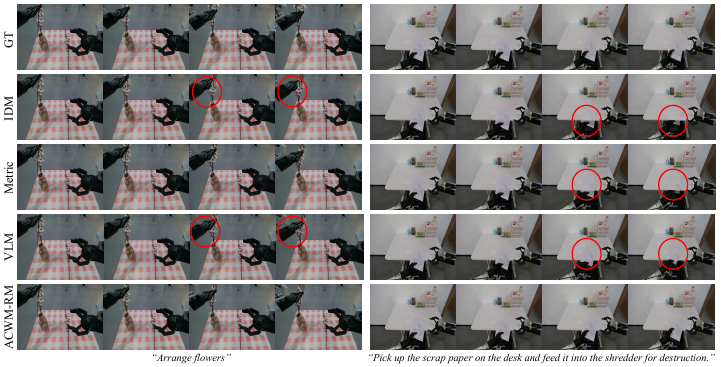}\par
\captionof{figure}{Qualitative comparison of generated videos with different reward signals.}
\label{fig:case}
\end{minipage}
\vspace{-12pt}
\end{table*}

\subsection{Flow-GRPO with Different Reward Signals}
\label{sec:reward_model_performance_for_Flow-GRPO}

We evaluate ACWMs trained with different rewards on $\mathrm{Agi_C}$ along the following dimensions.

\textbf{Visual quality.}
ACWM-RM achieves the best results across all five VQ metrics in Table~\ref{tab:flow-grpo-quality-consistency}, outperforming the full-reference metric-based reward.
Unlike this baseline, ACWM-RM requires no ground-truth videos for reward computation during RL.
Although its training preferences are derived from PSNR and LPIPS, the gains extend to SSIM, MSE, and FVD, which are not used for preference supervision.
In contrast, the IDM-based reward degrades all five metrics relative to Origin, suggesting that optimizing IDM-based AF alone is insufficient to preserve visual quality.

\textbf{Action-following fidelity.}
Table~\ref{tab:flow-grpo-action-consistency} shows that ACWM-RM is the only evaluated reward that improves nDTW over Origin, from 14.84 to 16.41.
The VLM-based and metric-based rewards yield no improvement, while the IDM-based reward reduces nDTW to 13.21 despite explicitly measuring consistency between inferred and conditioning actions.
Together with the AVM-Eval findings, these results suggest that strong action-mismatch discrimination on real videos does not necessarily translate into an effective RL reward for improving action-following fidelity.


\textbf{Physical consistency.}
Although ACWM-RM does not explicitly reward physical consistency, the model optimized with it achieves scores comparable to Origin across all three criteria in Table~\ref{tab:flow-grpo-quality-consistency}.
By contrast, the VLM-based reward yields a lower overall physical-consistency score than Origin (3.86 vs.\ 4.00), despite directly targeting these criteria.
One possible explanation is disagreement between the VLMs used for reward computation and evaluation: optimizing scores assigned by the reward VLM may not improve physical consistency as judged by the evaluation VLM.

\subsection{Ablation Study}
\label{sec:ablation_study}

\textbf{Score head architecture.}
We freeze the pretrained predictor and optimize the remaining head parameters on AVM-Pref.
Starting from the complete head (Frozen Predictor), we cumulatively replace learned patch weighting with uniform averaging (w/o weight MLP), replace comparison-vector-based scoring with cosine similarity between projected features (w/o comparison), and remove the projection (w/o projection).
With the predictor frozen, the complete head improves video-mismatch accuracy from 55.0 to 79.0.
Most of this gain comes from learned comparison-based scoring (59.4 to 77.7), while patch weighting provides a further improvement to 79.0 (Table~\ref{tab:ablation-agibot}).

\textbf{Complementary preference supervision.}
Table~\ref{tab:ablation-agibot} shows the complementary roles of both pair types (Sec.~\ref{action-video-preference-dataset}).
Training only on action-mismatch pairs (w/o video-pair) matches Ours at 91.6 on real-video action-mismatch pairs but reaches 71.8 on generated-video pairs.
Training only on video-mismatch pairs (w/o action-pair) achieves 82.5 on real-video action-mismatch pairs and 77.3 on generated-video pairs.
Combining both achieves 91.6 and 88.6, respectively, suggesting that action-mismatch pairs supervise AF, while video-mismatch pairs help transfer it to generated videos.

\begingroup
\setlength{\intextsep}{4pt}
\begin{table}[H]
\centering
\caption{In-domain ablation results for four aspects of ACWM-RM on AVM-Eval.}
\label{tab:ablation-agibot}
\begin{adjustbox}{width=\referenceTableWidth}
\begin{NiceTabular}{lcc!{\vrule}cc!{\vrule}c|cc!{\vrule}cc!{\vrule}c}
\toprule
\Block[m]{3-1}{Model} & \multicolumn{5}{c|}{Action-mismatch Pair} & \multicolumn{5}{c}{Video-mismatch Pair} \\
\cmidrule(lr){2-6}\cmidrule(lr){7-11}
 & \multicolumn{2}{c!{\vrule}}{$\mathrm{Agi_A^{test}}$} & \multicolumn{2}{c!{\vrule}}{$\mathrm{Agi_B}$} & \Block[m]{2-1}{Overall} & \multicolumn{2}{c!{\vrule}}{$\mathrm{Agi_A^{test}}$} & \multicolumn{2}{c!{\vrule}}{$\mathrm{Agi_B}$} & \Block[m]{2-1}{Overall} \\
\cmidrule{2-5}\cmidrule{7-10}
 & N-P & S-P & N-P & S-P & & Intra & Cross & Intra & Cross & \\
\midrule
Frozen Predictor & 97.0 & 78.8 & 95.0 & 76.5 & 86.8 & \underline{80.0} & \underline{84.3} & \underline{77.2} & \underline{74.5} & \underline{79.0} \\
\hspace*{0.5em}w/o weight MLP & 96.9 & 79.2 & 95.4 & 76.2 & 86.9 & 77.9 & 82.9 & 75.8 & 74.1 & 77.7 \\
\hspace*{0.5em}w/o comparison & 96.2 & 76.5 & 94.2 & 75.2 & 85.5 & 57.8 & 61.8 & 58.9 & 59.1 & 59.4 \\
\hspace*{0.5em}w/o projection & 96.0 & 76.4 & 93.9 & 75.1 & 85.4 & 53.7 & 55.7 & 55.7 & 54.7 & 55.0 \\
\midrule
Fresh Predictor & 92.0 & 77.2 & 92.2 & 78.4 & 85.0 & 61.6 & 61.8 & 62.7 & 61.9 & 62.0 \\
JEPA-IDM & 95.1 & 84.1 & 93.1 & 81.7 & \underline{88.5} & 73.2 & 76.3 & 73.6 & 71.5 & 73.7 \\
\midrule
Ours & \textbf{97.9} & \textbf{86.1} & \textbf{97.7} & \underline{84.7} & \textbf{91.6} & \textbf{90.5} & \textbf{93.3} & \textbf{86.2} & \textbf{84.3} & \textbf{88.6} \\
\hspace*{0.5em}w/o action-pair & 91.5 & 75.8 & 89.4 & 73.1 & 82.5 & 78.2 & 81.4 & 76.2 & 73.4 & 77.3 \\
\hspace*{0.5em}w/o video-pair & \underline{97.7} & \underline{85.8} & \underline{97.4} & \textbf{85.5} & \textbf{91.6} & 71.9 & 73.9 & 70.6 & 70.8 & 71.8 \\
\bottomrule
\end{NiceTabular}
\end{adjustbox}
\end{table}

\begin{wraptable}{r}{0.50\textwidth}
\vspace{-4pt}
\centering\footnotesize
\setlength{\abovecaptionskip}{0pt}
\setlength{\belowcaptionskip}{4pt}
\makeatletter
\long\def\@makecaption#1#2{%
  \vskip\abovecaptionskip
  {\small\raggedright #1: #2\par}%
  \vskip\belowcaptionskip}
\makeatother
\caption{Out-of-domain ablation results for four aspects of ACWM-RM on AVM-Eval.}
\label{tab:ablation-ood}
\setlength{\tabcolsep}{2pt}
\renewcommand{\arraystretch}{1.05}
\begin{adjustbox}{width=\linewidth}
\begin{NiceTabular}{lcc!{\vrule}cc!{\vrule}cc!{\vrule}c}
\toprule
\Block[m]{2-1}{Model} & \multicolumn{2}{c!{\vrule}}{$\mathrm{Gen}$} & \multicolumn{2}{c!{\vrule}}{$\mathrm{Robo_T}$} & \multicolumn{2}{c!{\vrule}}{$\mathrm{Robo_D}$} & \Block[m]{2-1}{Overall} \\
\cmidrule{2-7}
 & N-P & S-P & N-P & S-P & N-P & S-P & \\
\midrule
Frozen Predictor & \textbf{72.8} & \textbf{61.3} & \underline{77.7} & \underline{71.7} & 49.0 & 46.8 & \underline{63.2} \\
\hspace*{0.5em}w/o weight MLP & 65.4 & 55.4 & 76.2 & 71.0 & 47.5 & 47.3 & 60.5 \\
\hspace*{0.5em}w/o comparison & 62.9 & 52.4 & 76.3 & 69.6 & 49.8 & 46.9 & 59.7 \\
\hspace*{0.5em}w/o projection & 62.4 & 51.4 & 75.3 & 69.5 & 50.2 & 46.3 & 59.2 \\
\midrule
Fresh Predictor & 50.5 & 46.5 & 67.0 & 63.1 & 64.9 & 53.2 & 57.5 \\
JEPA-IDM & 63.1 & 53.1 & 73.6 & 66.1 & \underline{66.1} & \underline{57.1} & 63.2 \\
\midrule
Ours & \underline{65.9} & \underline{55.5} & \textbf{77.8} & \textbf{72.4} & \textbf{66.2} & \textbf{60.9} & \textbf{66.5} \\
\hspace*{0.5em}w/o action-pair & 62.6 & 52.5 & 71.6 & 61.2 & 49.3 & 48.4 & 57.6 \\
\hspace*{0.5em}w/o video-pair & 54.8 & 49.7 & 76.8 & 71.6 & 48.0 & 45.9 & 57.8 \\
\bottomrule
\end{NiceTabular}
\end{adjustbox}
\end{wraptable}

\textbf{Dynamics pretraining and adaptation.}
We examine whether real-video dynamics pretraining and joint preference training improve AF assessment.
We construct two variants using the same score head and AVM-Pref: Fresh Predictor jointly trains a randomly initialized predictor and the head, while Frozen Predictor trains only the head with the pretrained predictor fixed.
Compared with Fresh Predictor, Ours improves action-mismatch accuracy from 85.0 to 91.6 and video-mismatch accuracy from 62.0 to 88.6 (Table~\ref{tab:ablation-agibot}).
These gains support the value of learning dynamics before preference training.
Compared with Frozen Predictor, Ours improves video-mismatch accuracy from 79.0 to 88.6 under the same preference supervision, showing that adapting the pretrained predictor during preference training further benefits AF assessment on generated videos.

\textbf{Comparison with an inverse-dynamics architecture.}
To assess whether ACWM-RM's advantage can be explained solely by exposure to preference data, we construct JEPA-IDM with the same frozen V-JEPA 2.1 encoder and predictor backbone, but use action prediction instead of visual-feature prediction.
Following the IDM reward formulation, JEPA-IDM uses action-prediction errors as rewards and undergoes action-prediction pretraining on real videos from $\mathrm{Agi_A^{train}}$, followed by preference training on AVM-Pref.
Implementation details are provided in Appendix~\ref{app:jepa-idm}.
As shown in Table~\ref{tab:ablation-agibot}, ACWM-RM outperforms JEPA-IDM by 3.1 points on action-mismatch pairs (91.6 vs.\ 88.5) and 14.9 points on video-mismatch pairs (88.6 vs.\ 73.7).
The larger gap on generated videos supports the benefit of ACWM-RM's feature prediction and scoring design, even when JEPA-IDM has the same backbone and training data and follows the same two-stage training paradigm.
\par
\endgroup

\section{Conclusion, Limitations, and Future Work}
\label{sec:conclusion}
Through AVM-Eval, we showed that strong action-mismatch discrimination on real videos does not ensure reliable AF assessment of generated videos.
To address this limitation while also assessing VQ, we introduced ACWM-RM, a reward model requiring no ground-truth videos for reward computation.
ACWM-RM outperforms IDMs on generated-video AF and, when used to fine-tune DreamDojo-2B, yields better VQ and AF than alternative reward signals.

\textbf{Limitations and future work.} 
ACWM-RM is trained to assess VQ and AF, but does not explicitly supervise physical consistency (PC). 
Although RL fine-tuning with ACWM-RM maintains PC scores close to those of the pre-RL model, the reward does not directly target physical violations such as interpenetration.
Moreover, downstream RL improvements have so far been demonstrated only for DreamDojo-2B on AgiBotWorld.
Future work will incorporate physical-consistency supervision and evaluate ACWM-RM for RL optimization across a broader range of ACWMs.

\clearpage
\subsection*{AI use statement}
We used generative AI tools to polish the prose and check the grammar of author-written text.
We also used vision-language models in our experiments: Qwen3-VL provided the VLM-based reward during Flow-GRPO, and GPT-5.6 sol evaluated the physical consistency of generated videos, as described in Sec.~\ref{sec:experimental_settings} and Appendix~\ref{app:prompts_and_criteria}.
We did not use generative AI to generate research ideas.
The authors checked each AI-polished passage to ensure that it preserved the intended technical meaning, and reviewed and tested all AI-assisted code.
We take responsibility for the final content of this work, including all AI-assisted text, code, and experimental results.

\subsection*{Reproducibility statement}
We describe dataset construction and evaluation protocols in Secs.~\ref{sec:avmb}, \ref{action-video-preference-dataset}, and~\ref{sec:experimental_settings}.
Training configurations and hyperparameters for ACWM-RM, the baseline IDMs, JEPA-IDM, and Flow-GRPO are provided in Appendices~\ref{app:reward-model-implementation}, \ref{app:idm-training-setups}, \ref{app:jepa-idm}, and~\ref{app:flow_grpo_details}, respectively.
Additional appendices detail action perturbations, preference-pair construction, evaluation prompts, and dataset statistics.

\bibliography{iclr2027_conference}
\bibliographystyle{iclr2027_conference}

\appendix
\section{Action Perturbations for Action-Mismatch Pairs}
\label{app:action_perturbation}
This appendix details how the two classes of action perturbations are performed.
Let a clip be $\mathbf{I}=(I_0,\ldots,I_{12})$ and the corresponding end-effector states be $\mathbf{s}=(s_0,\ldots,s_{12})$.  
The clip therefore contains 12 consecutive action transitions $\mathbf{a}=(a_1,\ldots,a_{12})$.  
For arm $r\in\{L,R\}$, state $s_t^r$ consists of a position $x_t^r\in\mathbb{R}^{3}$, an orientation quaternion $q_t^r$ in $(q_x,q_y,q_z,q_w)$ order, and a scalar gripper value $\gamma_t^r$ in degrees.
The action at transition $t$, which links $I_{t-1}$ to $I_t$, is
\begin{equation}
 a_t =
 [\Delta x_t^L,\Delta x_t^R,\Delta\omega_t^L,\Delta\omega_t^R,
  \Delta\gamma_t^L,\Delta\gamma_t^R]\in\mathbb{R}^{14},
 \qquad t=1,\ldots,12,
 \label{eq:local-action-layout}
\end{equation}
with
\begin{align}
 \Delta x_t^r &=x_t^r-x_{t-1}^r, &
 \Delta\omega_t^r&=\operatorname{axisangle}
       \!\left(q_t^r\otimes(q_{t-1}^r)^{-1}\right), &
 \Delta\gamma_t^r&=\gamma_t^r-\gamma_{t-1}^r .
 \label{eq:local-action-delta}
\end{align}

We refer to each of the six groups $\Delta x_t^r$, $\Delta\omega_t^r$ and $\Delta\gamma_t^r$ ($r\in\{L,R\}$) as a channel, so that every action carries the three channel types: translation, rotation and gripper.
Throughout, $t\in\{1,\ldots,12\}$ indexes the transitions of a clip and $j\in\{0,\ldots,13\}$ indexes the action dimensions of Eq.~\ref{eq:local-action-layout}, so that $a_{t,j}$ is a single scalar of $a_t$.
The clean action chunk is
$\mathbf{a}^{+}=(a_1,\ldots,a_{12})\in\mathbb{R}^{12\times14}$.  
Both mismatch classes retain the same video and use a modified action chunk $\mathbf{a}^{-}$ as the negative condition.
Appendix~\ref{app:predictor} writes the same three components as $\Delta p_t$, $\omega_t$ and $\Delta g_t$ after normalization, where the gripper is rescaled from degrees to $[0,1]$.

\subsection{Active Channel Identification}
\label{app:local-activity}

ACWMs predict a stationary arm well but a moving arm poorly, so their prediction error is concentrated in the channels in which an arm is moving. 
These are therefore the channels a reward model has to discriminate, and we restrict perturbations to them.

To identify these channels, we compute six Boolean activity flags, one per channel.
Each flag is set from the spread of the absolute trajectory over the 13 frames rather than summed per-step absolute changes, preventing small mean-reverting servo jitter from being mistaken for sustained motion.

For each arm $r$, compute
\begin{equation}
 \begin{aligned}
 &S_{\mathrm{trans}}^r
   =\max_{0\leq t,t'\leq12}\|x_t^r-x_{t'}^r\|_2,\\
 &S_{\mathrm{rot}}^r
   =\max_{0\leq t,t'\leq12}
      2\arccos\!\left(\left|\left\langle
      q_t^r,q_{t'}^r\right\rangle\right|\right),\\
 &S_{\mathrm{grip}}^r
   =\max_{0\leq t\leq12}\gamma_t^r-
     \min_{0\leq t\leq12}\gamma_t^r .
 \end{aligned}
 \label{eq:local-spreads}
\end{equation}
The first quantity is the diameter of the arm's 3D position trajectory, the second is the largest geodesic orientation difference, and the third is the peak-to-peak gripper excursion.
The resulting activity flags are
\begin{equation}
 \begin{aligned}
 &\mathrm{active}_{\mathrm{trans}}^r
   \Longleftrightarrow S_{\mathrm{trans}}^r>\tau_{\mathrm{trans}}^r,\\
 &\mathrm{active}_{\mathrm{rot}}^r
   \Longleftrightarrow S_{\mathrm{rot}}^r>\tau_{\mathrm{rot}}^r,\\
 &\mathrm{active}_{\mathrm{grip}}^r
   \Longleftrightarrow S_{\mathrm{grip}}^r>\tau_{\mathrm{grip}}.
 \end{aligned}
 \label{eq:local-activity-flags}
\end{equation}
\begin{figure}[t]
\begin{center}
\includegraphics[width=\linewidth]{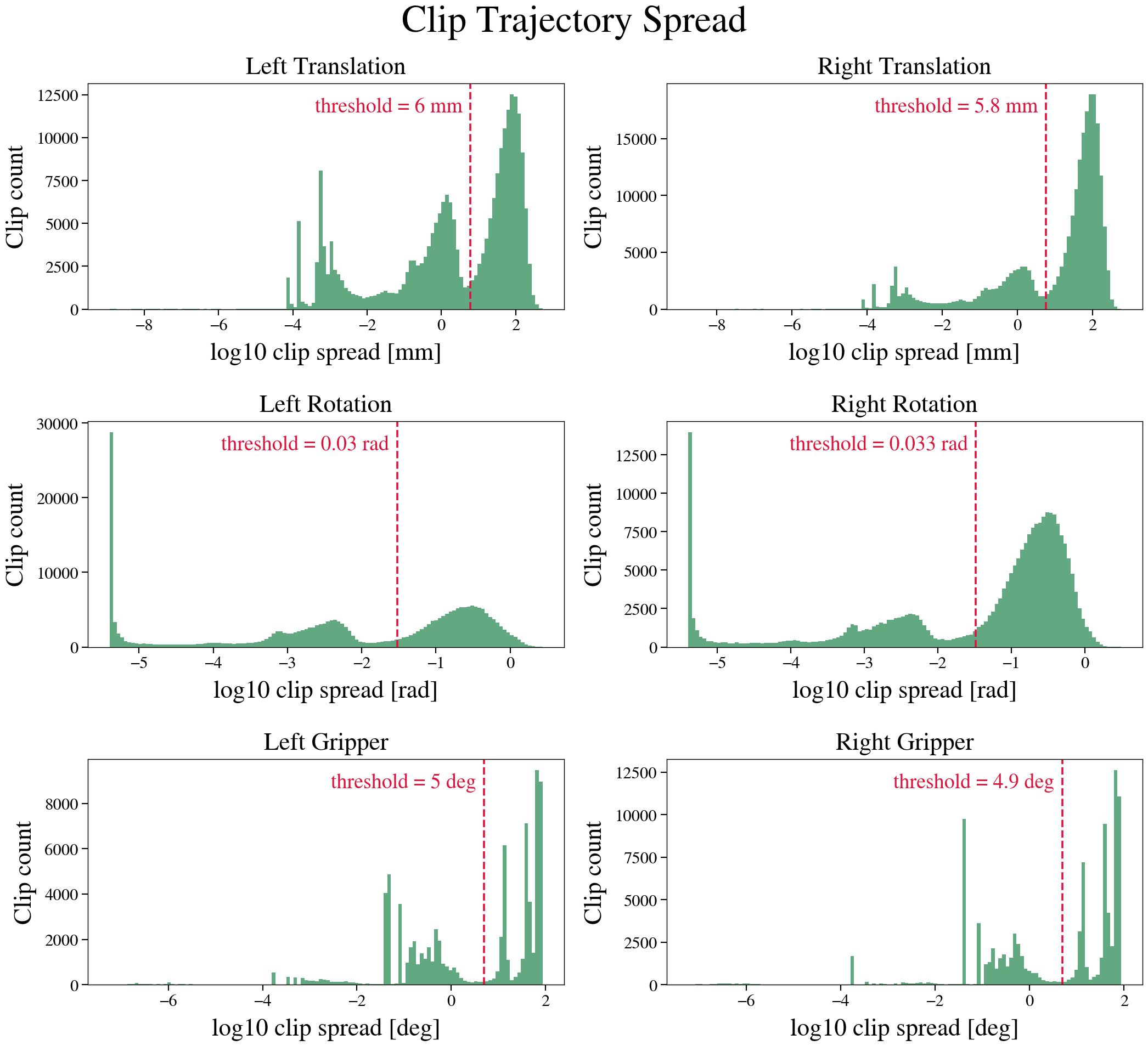}
\end{center}
\caption{Clip trajectory spread.
Each panel histograms the $\log_{10}$ spread statistic of Eq.~\ref{eq:local-spreads} over the extracted clips, with one column per arm and one row per channel type.
The red dashed line marks the threshold calibrated for that channel, placed slightly to the right of the valley separating them.}
\label{fig:clip-spread}
\end{figure}

We calibrate the thresholds offline by collecting a large number of clips and histogramming each spread statistic separately for the left and right arm, as shown in Figure~\ref{fig:clip-spread}.
Every histogram is bimodal, with small spreads forming the static and servo-noise population and larger spreads forming the genuine-motion population.
Each threshold is placed slightly to the right of the valley separating the two populations, and channels above it are marked active. 
Biasing the threshold in this direction trades false positives for false negatives: we accept that a few low-amplitude moving channels are missed in order to keep static channels out of the active set.
Each arm has its own valley, so the calibration yields six separate values, namely 6\,mm and 5.8\,mm for left and right translation, 0.03\,rad and 0.033\,rad for rotation, and 5 and 4.9 degrees for the gripper, held fixed during pair generation.

The flags determine the eligible perturbations.  
For example, if only the left-translation and right-gripper channels are active, perturbations for that action chunk are drawn only from the left-translation and right-gripper operators.
Clips with no active channel are discarded.

\subsection{Numerical Action Perturbations}
\label{app:local-numerical}

Numerical action perturbations modify $\mathbf{a}^{+}$ using operators drawn from a perturbation pool whose membership is determined by the six activity flags.
We first select the temporal windows within the action chunk to be edited, then assemble the perturbation pool, then apply the drawn perturbations channel by channel, and finally check whether the negative differs sufficiently from the positive.

\paragraph{Temporal Window Sampling.}

Perturbing every action would make the negative wrong everywhere, which is both easy to detect and unrepresentative of the localized errors an ACWM actually makes. 
We therefore confine each edit to a few windows of consecutive actions and leave the remaining actions clean. 
In addition,  many different combinations of windows and perturbation operators remain available, so a single action chunk can yield many distinct negatives.

The windows themselves are sampled in one of two ways. 
With probability 0.3, the whole chunk is taken as a single window. 
Retaining these fully perturbed negatives keeps a coarse difficulty level in the mix.
Otherwise, we sample one to three windows, each covering one to seven consecutive actions, and require neighboring windows to be separated by at least one unedited action so that the edited segments stay distinguishable. 
If the sampled windows cannot all be placed within the 12 actions under this constraint, we drop windows until the remaining ones fit.
In either case, the outcome is a set of windows $\mathcal{W}=\{[w_k^{\rm lo},w_k^{\rm hi}]\}$ marking the actions to be edited, where $k$ indexes the windows and $w_k^{\rm lo}$, $w_k^{\rm hi}$ are the first and last actions of window $k$. The actions outside them stay clean.

\paragraph{Perturbation Pool Construction and Sampling.}

To enlarge the range of perturbation types, we build a perturbation pool for each action chunk to be perturbed.
For every arm with an active translation flag, we draw one translation perturbation with equal probability from \textit{Magnitude}, \textit{Direction}, and \textit{Magnitude \& Direction}, and add it to the pool.
For every arm with an active rotation flag, we draw one rotation perturbation with equal probability from \textit{Orientation Direction}, \textit{Orientation Magnitude}, and \textit{Orientation Magnitude \& Direction}, and add it to the pool.
For every active gripper flag, we add one \textit{Gripper Perturbation} operator to the pool.
The pool further holds a global \textit{Gaussian Perturbation} whenever at least one channel is active, together with a global \textit{Temporal Order Permutation} operator.
Each perturbation in the pool is then selected independently with probability 0.5, and if no perturbation is selected, we select one uniformly from the pool, so that every negative carries at least one edit.
We initialize $\mathbf{a}^{-}=\mathbf{a}^{+}$ before applying the selected perturbations sequentially.
All selected perturbations are applied one after another to all sampled windows to produce a single $\mathbf{a}^{-}$, with the structured perturbations writing their own action dimensions, Gaussian noise added on top, and order permutation applied last.

\paragraph{Perturbation Types.} We describe each perturbation type below:

\begin{enumerate}[label=\arabic*.,leftmargin=1.6em,itemsep=4pt,topsep=4pt]

\item \textbf{Translation Perturbations.}
Translation perturbations act on the translation deltas $\Delta x_t^{r,+}$ and write the perturbed deltas $\Delta x_t^{r,-}$.
They comprise the following three types of perturbations.

\begin{enumerate}[label=(\arabic*),leftmargin=1.8em,itemsep=2pt,topsep=2pt]
 \item \textit{Magnitude.} For each window, draw
 $\mu\sim U[0.4,0.8]$ or $\mu\sim U[1.2,1.6]$ and set
 \begin{equation}
   \Delta x_t^{r,-}=\mu\,\Delta x_t^{r,+},\qquad t\in[w_k^{\rm lo},w_k^{\rm hi}].
   \label{eq:local-scale}
 \end{equation}
 The direction is preserved, but the motion is under- or over-executed.

 \item \textit{Direction.} Compute the arm's net translation
 $m^r=\sum_t \Delta x_t^{r,+}$.
 When $\|m^r\|_2\geq10^{-6}$, sample a random unit axis $v^r\perp m^r$ and draw a signed
 $\eta\sim U[0.15,0.5]$ radians per window.
 Then
 \begin{equation}
   \Delta x_t^{r,-}=\mathrm{Rot}(v^r,\eta)\,\Delta x_t^{r,+},
   \label{eq:local-direction}
 \end{equation}
 where $\mathrm{Rot}$ is Rodrigues rotation.
 Each action keeps its length but points in a different direction.

 \item \textit{Magnitude \& Direction.} Apply both operations:
 \begin{equation}
   \Delta x_t^{r,-}=\mu\,\mathrm{Rot}(v^r,\eta)\,\Delta x_t^{r,+}.
   \label{eq:local-scale-direction}
 \end{equation}
 For an active translation channel with $\|m^r\|_2<10^{-6}$, both \textit{Direction} and \textit{Magnitude \& Direction} fall back to \textit{Magnitude}.
\end{enumerate}

\item \textbf{Rotation Perturbations.}
Rotation perturbations act on the rotation deltas $\Delta\omega_t^{r,+}$, whose norm is the turn amount and whose direction is the turn axis.
The three types are:

\begin{enumerate}[label=(\arabic*),leftmargin=1.8em,itemsep=2pt,topsep=2pt]
 \item \textit{Orientation Direction.} For each arm with an active rotation channel, define the aggregate rotation direction
 $m_{\mathrm{rot}}^r=\sum_{t=1}^{12}\Delta\omega_t^{r,+}$ over the entire clip.
 When $\|m_{\mathrm{rot}}^r\|_2\geq10^{-6}$, sample a random unit axis
 $v^r\perp m_{\mathrm{rot}}^r$ and apply the following perturbation only within the sampled windows:
 \begin{equation}
  \Delta\omega_t^{r,-}=\mathrm{Rot}(v^r,\eta)\,\Delta\omega_t^{r,+},\qquad
  \eta\sim U[0.3,0.8].
  \label{eq:local-rot-heading}
 \end{equation}
 When an axis is available, the turn amount is preserved, but the turn axis is changed.

 \item \textit{Orientation Magnitude.} Draw
 $\mu\sim U[0.5,0.8]$ or $\mu\sim U[1.2,1.5]$ and set
 \begin{equation}
  \Delta\omega_t^{r,-}=\mu\,\Delta\omega_t^{r,+}.
  \label{eq:local-rot-magnitude}
 \end{equation}
 The axis is preserved, but the turn amount is too small or too large.

 \item \textit{Orientation Magnitude \& Direction.} Apply both the axis change and the magnitude scale:
 \begin{equation}
  \Delta\omega_t^{r,-}=\mu\,\mathrm{Rot}(v^r,\eta)\,\Delta\omega_t^{r,+}.
  \label{eq:local-rot-magnitude-direction}
 \end{equation}
 For an active rotation channel with $\|m_{\mathrm{rot}}^r\|_2<10^{-6}$, both \textit{Orientation Direction} and \textit{Orientation Magnitude \& Direction} fall back to \textit{Orientation Magnitude}.
\end{enumerate}

\item \textbf{Gripper, Gaussian, and Temporal Order Perturbations.}

\begin{enumerate}[label=(\arabic*),leftmargin=1.8em,itemsep=2pt,topsep=2pt]
\item \textit{Gripper Perturbation.} Draw
$\mu\sim U[0.3,0.7]$ or $\mu\sim U[1.3,2.0]$ per window and set
\begin{equation}
 \Delta\gamma_t^{r,-}=\mu'\,\Delta\gamma_t^{r,+},\qquad
 \mu'=\begin{cases}-\mu,&\text{with probability }0.25,\\
 \mu,&\text{otherwise.}\end{cases}
 \label{eq:local-gripper}
\end{equation}
A sign flip changes closing into opening or opening into closing.  

\item \textit{Gaussian Perturbation.}
Let $a_{t,j}^{-}$ denote the current action value after any preceding structured perturbations.
We draw $\sigma_w\sim U[0.25,2.5]$ for each window and add independent noise to every active dimension:
\begin{equation}
 a_{t,j}^{-}\leftarrow a_{t,j}^{-}+\epsilon_{t,j},\qquad
 \epsilon_{t,j}\sim\mathcal{N}\!\left(0,
   (\sigma_w\,\sigma_{\mathrm{act},j})^2\right).
 \label{eq:local-gaussian}
\end{equation}
Here $\sigma_{\mathrm{act},j}$ is the empirical standard deviation of dimension $j$ on clips whose corresponding channel is active.

\item \textit{Temporal Order Permutation.} Within each window containing at least two actions, reversal, random shuffling, or a cyclic roll reorders only active dimensions; inactive dimensions remain unchanged.
This preserves the active values after any preceding perturbations and, when applied alone, changes only their temporal order.
Permutations that leave the window unchanged are discarded and resampled.
\end{enumerate}

\end{enumerate}

\paragraph{Filtering Degenerate Negatives.}
Applying the selected perturbations does not by itself guarantee a usable negative.
A perturbation drawn near the mild end of its range, or applied to a window in which the arm barely moves, can leave $\mathbf{a}^{-}$ almost identical to $\mathbf{a}^{+}$ while the pair is still labeled a mismatch.
We therefore keep a pair only when it is far enough from the positive.

Measuring this distance requires care on two counts.
The 14 action dimensions of Eq.~\ref{eq:local-action-layout} carry different units, so the differences $a^{+}_{t,j}-a^{-}_{t,j}$ cannot be summed across dimensions.
And the perturbations apply only to the active channels inside the sampled windows, leaving most of these differences at zero, so an average taken over all positions would be diluted by the clean part.
We address the first by dividing $a^{+}_{t,j}-a^{-}_{t,j}$ by $\sigma_j$, the empirical standard deviation of dimension $j$ over all transitions of the extracted clips,
\begin{equation}
\begin{aligned}
\sigma = \big(
&\underbrace{0.00316,\ 0.00397,\ 0.00372}_{\text{left trans.\ [m]}},\
\underbrace{0.00335,\ 0.00504,\ 0.00417}_{\text{right trans.\ [m]}},\\
&\underbrace{0.01535,\ 0.01566,\ 0.01552}_{\text{left rot.\ [rad]}},\
\underbrace{0.01622,\ 0.01437,\ 0.01608}_{\text{right rot.\ [rad]}},\
\underbrace{3.950,\ 4.287}_{\text{grippers [deg]}}
\big),
\end{aligned}
\label{eq:local-sigma}
\end{equation}
measured at the 8\,fps clips.
We address the second by averaging only over the positions $(t,j)$ at which $a^{+}_{t,j}\neq a^{-}_{t,j}$, so that the score reflects the average displacement of each edited action rather than the number of actions edited.
Dividing per dimension by $\sigma$, the resulting distance is
\begin{equation}
 \mathrm{dist}(\mathbf{a}^{+},\mathbf{a}^{-})=
 \frac{\left\|(\mathbf{a}^{+}-\mathbf{a}^{-})/\sigma\right\|_F}
      {\sqrt{\#\{(t,j):a^{+}_{t,j}\neq a^{-}_{t,j}\}}}.
 \label{eq:local-distance}
\end{equation}
A pair is kept only if $\mathbf{a}^{-}$ differs from $\mathbf{a}^{+}$ and $\mathrm{dist}(\mathbf{a}^{+},\mathbf{a}^{-})\geq0.12$.
We set this floor from the activity thresholds of Sec.~\ref{app:local-activity}.
A channel only counts as active once its 13-frame trajectory spread exceeds $\tau^r$, which corresponds to an average per-transition motion of $\tau^r/12$, or $\tau^r/(12\sigma)$ once expressed in the units of Eq.~\ref{eq:local-distance}.
Across the six channels, this quantity ranges from $0.095$ to $0.177$ with a median of $0.126$, and we round it to $0.12$.
Negatives whose average per-transition change falls below this level are therefore discarded.
The floor deserves particular attention when \textit{Temporal Order Permutation} is applied alone: although it preserves the positive action values, changing their temporal order does not guarantee a sufficiently large action mismatch.
When the reordered window moves at near-constant velocity, however, those actions return to almost the same places, so the distance can still fall below the floor and the pair is discarded.
A rejected $\mathbf{a}^{-}$ triggers a fresh draw of the windows, the perturbation pool, and all magnitudes, for at most five attempts, after which the clip yields no pair and is discarded.

\subsection{Semantic Action Perturbations}
\label{app:local-semantic}

Semantic action perturbations differ from numerical action perturbations in what they edit and where the edits are applied.
The semantic procedure never perturbs $\mathbf{a}^{+}$ directly, but edits the absolute state sequence $\mathbf{s}$ and obtains $\mathbf{a}^{-}$ by differencing the edited states through Eq.~\ref{eq:local-action-delta}.
The perturbation is anchored at a grasp, release, or mid-motion event rather than at an individual action, so the error corresponds to a semantic-level mistake.
These events can be identified from the gripper events: a gripper closing marks a grasp, a gripper opening marks a release, and the interval between a closing and the following opening marks mid-motion drift or freeze.
We therefore locate these events and anchor the semantic perturbations to them.

Applying perturbation at a gripper event requires more context than a single clip provides.
We therefore subsample the episode to 8\,fps, detect the events and apply the perturbation over this whole subsampled episode, and crop the 13-frame clip afterward.

\paragraph{Gripper Event Detection.}
We first describe how gripper closing and opening events are extracted from an episode.
Let $\gamma_{\max}^r$ and $\gamma_{\min}^r$ be the largest and smallest values it takes over the whole episode, $\gamma_t^r$ be the gripper opening of arm $r$, and $\rho^r=\gamma_{\max}^r-\gamma_{\min}^r$ be its range.
If $\rho^r<5$ degrees, the gripper never opens or closes appreciably.
Otherwise set
\begin{equation}
 \gamma_{\rm lo}^r=\gamma_{\min}^r+0.30\rho^r,\qquad
 \gamma_{\rm hi}^r=\gamma_{\min}^r+0.70\rho^r.
 \label{eq:local-hysteresis}
\end{equation}
Here $\gamma_{\rm lo}^r$ and $\gamma_{\rm hi}^r$ are the closing and opening thresholds for arm $r$, placed at 30\% and 70\% of that arm's observed gripper range, so that the band between them spans the middle 40\%.
A single pass over the episode gives every frame an open or closed flag.
Specifically, a frame is labeled close when its $\gamma^r$ falls below $\gamma_{\rm lo}^r$ and open when it rises above $\gamma_{\rm hi}^r$, while frames lying between the two thresholds are left unlabeled.
Whenever two consecutive labeled frames carry opposite labels, the interval between them is marked as an event, which is a gripper close event when the labels run from open to close and a gripper open event in the reverse case.
Because the thresholds are 40\% apart, fluctuations near one of them cannot produce an event.
Then, for the mid-motion events, we scan each arm's $\gamma^r$ over the episode and mark a maximal interval as mid-motion if $\gamma^r$ stays below $\gamma_{\rm lo}^r$ throughout while the end-effector displacement across the interval exceeds $\tau_{\mathrm{trans}}^r$. 
(Intervals shorter than 25 frames are discarded.)
Finally, the detected three events are used by the three semantic perturbations described below.

\paragraph{Perturbation Types.} We describe each perturbation type below:

\begin{enumerate}[label=\arabic*.,leftmargin=1.6em,itemsep=4pt,topsep=4pt]

\item \textbf{Position offset.}
This type covers \textit{grasp offset}, anchored at a gripper close event, and \textit{place offset}, anchored at a gripper open event.
The two variants use the same construction, so we describe only one of them here.
Let $b$ be the first frame of the event interval, at which the gripper starts to change state, and $r$ be the arm on which it was detected.
We sample a window length $L$ from 10 to 25 frames and set $w_{\rm hi}=b$ and $w_{\rm lo}=\max(b-L,0)$, so the window covers the motion leading up to the event rather than the gripper transition itself.
We then take the arm's net displacement over the window, $m^r=x_{w_{\rm hi}}^{r,+}-x_{w_{\rm lo}}^{r,+}$, sample a random unit direction $v^r\perp m^r$ as in Eq.~\ref{eq:local-direction}, and draw an amplitude $\delta\sim U[0.005,0.050]$.
The arm's positions are shifted along $v^r$ by an offset that grows linearly over the window and is then held to the end of the episode, while frames before $w_{\rm lo}$ stay unchanged,
\begin{equation}
x_t^{r,-}=x_t^{r,+}
+\frac{t-w_{\rm lo}}{w_{\rm hi}-w_{\rm lo}}\,\delta v^r,
\quad t\in[w_{\rm lo},w_{\rm hi}],
\qquad
x_t^{r,-}=x_t^{r,+}+\delta v^r,\quad t>w_{\rm hi}.
\label{eq:local-semantic-offset}
\end{equation}
By the frame at which the gripper begins to close or open, the arm has already been displaced by the full $\delta$, so the grasp or release is performed at the wrong position.

\item \textbf{Gripper timing.}
This type covers four variants, obtained by pairing either gripper event with either direction of shift: \textit{early close}, \textit{late close}, \textit{early open}, and \textit{late open}.
Let $[b,e]$ denote the event interval and $r$ the arm on which it was detected.
We first sample a shift magnitude $d$ from 3 to 16 and set $\zeta=+d$ for a late variant and $\zeta=-d$ for an early one, which moves the event to $[b+\zeta,e+\zeta]$.
The gripper values on $[b,e]$ are copied to $[b+\zeta,e+\zeta]$, preserving the event's gripper-value profile while shifting it in time.
The gap between the original and the shifted interval is filled with a constant gripper value.
When the event is delayed, the gap lies before the shifted interval and is filled with $\gamma_b^{r,+}$, the gripper value at the first index of the original interval.
When the event is advanced, the gap lies after the shifted interval and is filled with $\gamma_e^{r,+}$, the gripper value at the last index of the original interval.
For a late variant ($\zeta=+d$),
\begin{equation}
\gamma_t^{r,-}=
\begin{cases}
    \gamma_{t-\zeta}^{r,+}, & b+\zeta\leq t\leq e+\zeta,\\
    \gamma_{b}^{r,+},       & b\leq t<b+\zeta,\\
    \gamma_t^{r,+},         & \text{otherwise},
\end{cases}
\label{eq:timing-late}
\end{equation}
and for an early variant ($\zeta=-d$),
\begin{equation}
\gamma_t^{r,-}=
\begin{cases}
    \gamma_{t-\zeta}^{r,+}, & b+\zeta\leq t\leq e+\zeta,\\
    \gamma_{e}^{r,+},       & e+\zeta<t\leq e,\\
    \gamma_t^{r,+},         & \text{otherwise}.
\end{cases}
\label{eq:timing-early}
\end{equation}
The edit spans the original and shifted intervals and any gap between them; gripper states outside this span remain unchanged.
When $d$ is smaller than the duration of the event, the two intervals overlap, and the second case of each equation covers only the part of the original interval that the shifted one does not.
The constant values used to fill the gap are taken from the corresponding boundaries of the original event interval.
The copied event profile is preserved, but continuity is not guaranteed at the joins with the unedited gripper trajectory.
These variants change grasp or release timing: the gripper closes before or after the object is in place, or opens before or after the arm has reached the target.

\item \textbf{Mid-motion drift or freeze.}
This type covers \textit{mid drift} and \textit{mid freeze}, both anchored at a mid-motion event.
Let $[b,e]$ be the event interval and $r$ be the arm on which it was detected.
We set $w_{\rm lo}=b$ and $w_{\rm hi}=e$, so the window covers the transport motion itself.
With probability 0.5 we apply the ramped offset of Eq.~\ref{eq:local-semantic-offset} to this window, with $m^r$, $v^r$ and $\delta$ obtained as before, so the arm drifts sideways while it carries the object.
With probability 0.5 we instead freeze the arm at the position it holds at $w_{\rm lo}$,
\begin{equation}
  x_t^{r,-}=
  \begin{cases}
    x_{w_{\rm lo}}^{r,+}, & w_{\rm lo}\leq t\leq w_{\rm hi},\\
    x_t^{r,+}-m^r,        & t>w_{\rm hi},\\
    x_t^{r,+},            & \text{otherwise},
  \end{cases}
  \qquad
  m^r=x_{w_{\rm hi}}^{r,+}-x_{w_{\rm lo}}^{r,+}.
  \label{eq:local-semantic-freeze}
\end{equation}
After the window, the arm moves exactly as in the original trajectory, but from a position that lags it by $m^r$, the net displacement it should have made while frozen.
Both variants edit translation only, leaving orientation and gripper values unchanged.
The arm therefore fails to carry the object along the demonstrated path, either drifting off course while transporting it or not transporting it at all.

\end{enumerate}

\paragraph{Pair Construction and Filtering.}

After cropping the 13-frame clip, we take the difference between the clean and the edited state sequences through Eq.~\ref{eq:local-action-delta} to obtain the two action chunks $\mathbf{a}^{+}$ and $\mathbf{a}^{-}$.
We keep the pair only if at least one of the 12 actions differs between them and the normalized distance of Eq.~\ref{eq:local-distance} is at least 0.12.
The first condition ensures that the crop overlaps the edited part of the episode, while the actions outside it stay clean and serve as context.
The resulting training pair is $(\mathbf{I},\mathbf{a}^{+})\succ(\mathbf{I},\mathbf{a}^{-}).$

\begin{figure}[t]
    \centering
    \includegraphics[width=\linewidth]{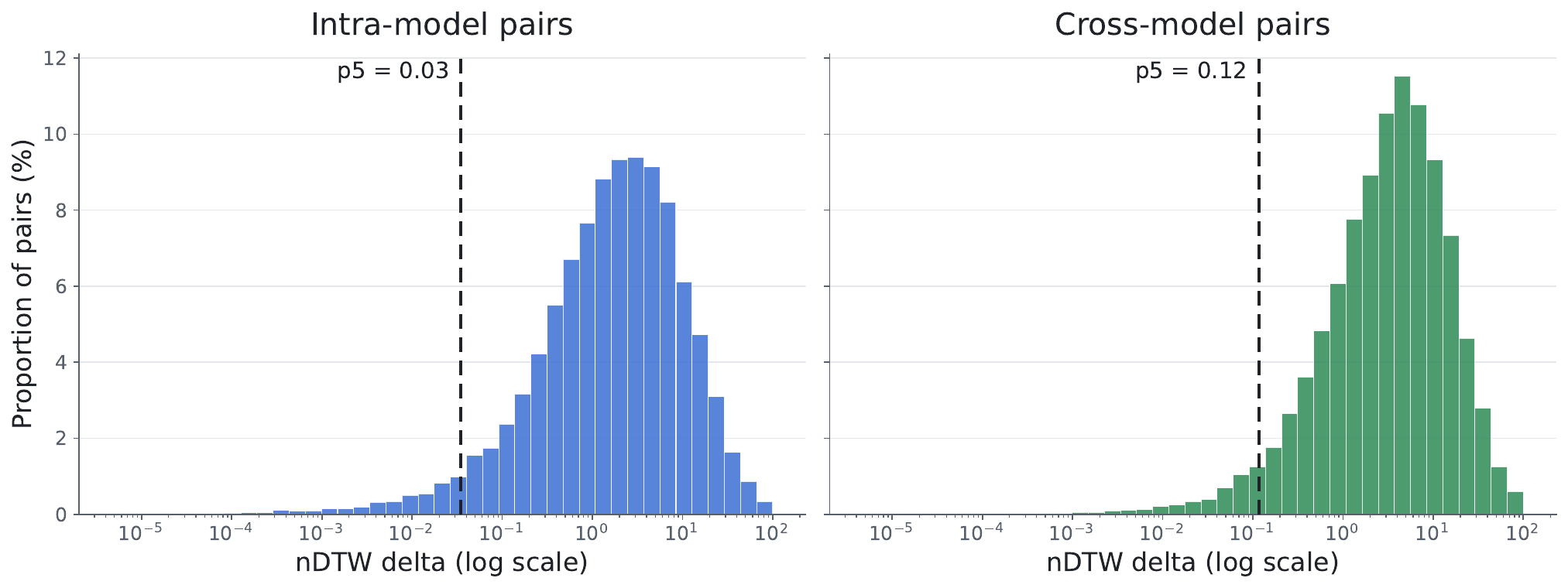}
    \caption{Distributions of absolute nDTW score differences ($\times 100$) for intra-model (left) and cross-model (right) candidate pairs from 5{,}760 rollouts conditioned on clips from $\mathrm{Agi_A^{test}}$ and $\mathrm{Agi_B}$. The horizontal axes use a logarithmic scale. Black vertical lines mark the respective 5th-percentile thresholds, 0.03 and 0.12; pairs below the corresponding threshold are discarded.}
    \label{fig:ndtw-pair-gap}
\end{figure}

\begin{figure}[t]
    \centering
    \includegraphics[width=\linewidth]{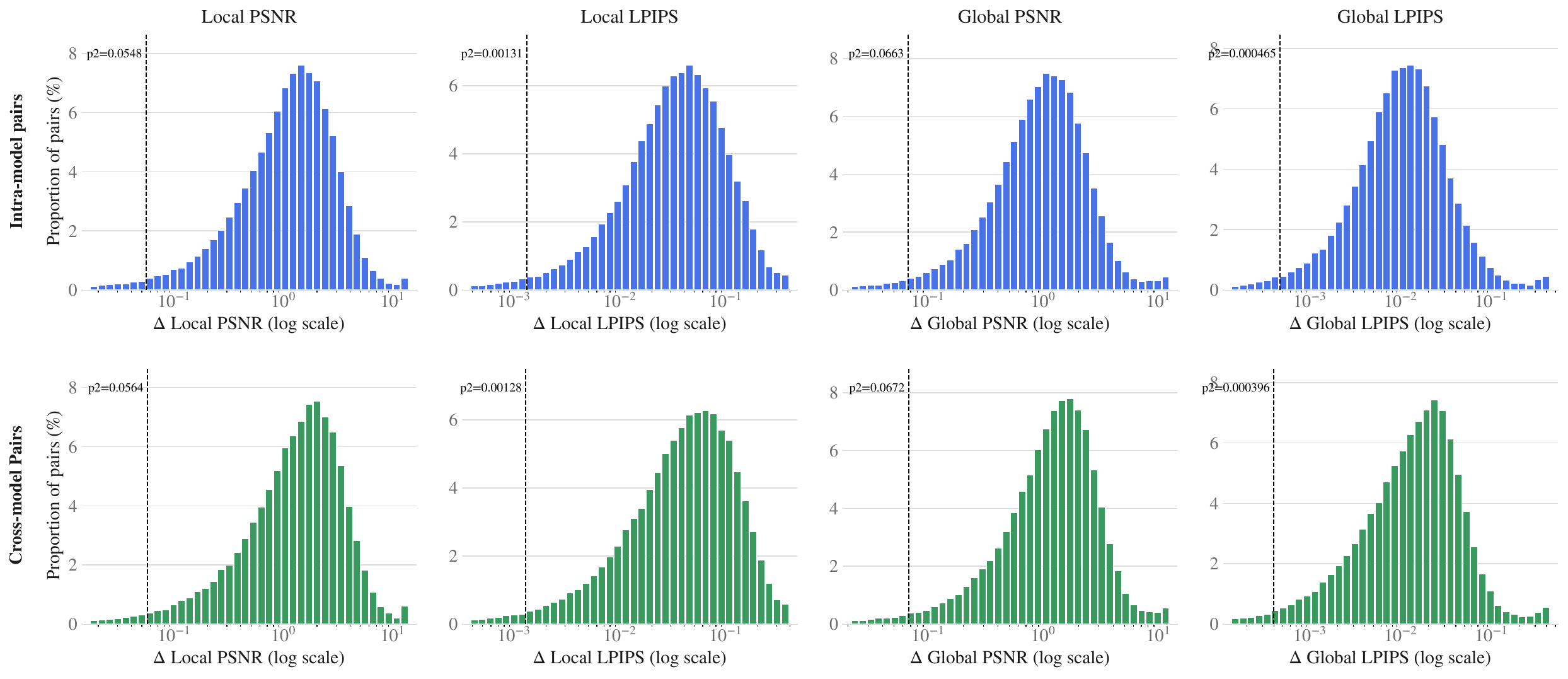}
    \caption{Visual-quality gap distributions for intra-model (top) and cross-model (bottom) candidate pairs from AgiBotWorld. Black dashed lines mark the 2nd-percentile thresholds.}
    \label{fig:visual-quality-gap}
\end{figure}

\section{Video-mismatch pair filtering for AVM-Eval}
\label{app:ndtw-pair-filtering}

\subsection{nDTW-Based Pair Filtering}
We compute nDTW scores using the official implementation of WorldArena~\citep{shang2026worldarena} and filter candidate video pairs with small score differences.
The threshold statistics are collected from rollouts conditioned on 240 input clips selected from $\mathrm{Agi_A^{test}}$ and $\mathrm{Agi_B}$, covering 30 tasks and 120 episodes, with eight clips per task.
For each clip, DreamDojo-14B, DreamDojo-2B, GE-Sim-2B, and OSCAR-2B each generate six videos under the same initial frame and action chunk, giving $240\times4\times6=5{,}760$ rollouts.

For each input clip, we enumerate all $\binom{6}{2}=15$ pairs within each model and all $6\times6=36$ pairs for each of the $\binom{4}{2}=6$ model combinations.
This gives 14{,}400 intra-model and 51{,}840 cross-model candidate pairs for the threshold statistics.
Let $s_i$ and $s_j$ denote the nDTW scores ($\times 100$) of two candidate videos relative to the same ground-truth video.
We compute their absolute score difference as
\begin{equation}
    \Delta_{ij}=|s_i-s_j|.
    \label{eq:ndtw-pair-gap}
\end{equation}
We pool these differences across input clips separately for intra-model and cross-model pairs, and set each filtering threshold to the 5th percentile of its corresponding distribution.
As shown in Figure~\ref{fig:ndtw-pair-gap}, the resulting thresholds are $\tau_{\mathrm{intra}}=0.03$ and $\tau_{\mathrm{cross}}=0.12$.
We retain an intra-model pair only when $\Delta_{ij}\geq\tau_{\mathrm{intra}}$ and a cross-model pair only when $\Delta_{ij}\geq\tau_{\mathrm{cross}}$, assigning the video with the higher score as preferred.
This empirical filtering rule excludes approximately the lowest 5\% of pairwise score differences in each distribution, removing candidate pairs with similar nDTW scores.

\subsection{Visual Quality Control}
\label{app:visual-quality-control}
To limit the influence of visual quality on AF evaluation, we further filter the pairs retained above using global and local PSNR and LPIPS, computed against the shared ground-truth video (Appendix~\ref{app:video-mismatch-pairs-for-AVM-Pre}).
To calibrate the thresholds, we construct candidate pairs from AgiBotWorld and order each pair by the quality metric being considered.
For each metric $m$ and pair type $c$ (intra-model or cross-model), we set $\tau_{m,c}$ to the 2nd percentile of the resulting positive quality gaps (Figure~\ref{fig:visual-quality-gap}).

For the pairs retained by nDTW filtering, we keep their preference labels fixed and define the signed quality gap $\delta_m$ as positive-minus-negative for PSNR and negative-minus-positive for LPIPS.
We retain a pair only if $|\delta_m| \leq \tau_{m,c}$ for all four metrics.
This ensures that the higher-nDTW and lower-nDTW videos have comparable visual quality within the empirical tolerances.

\section{Video-Mismatch Pair Construction for AVM-Pref}
\label{app:video-mismatch-pairs-for-AVM-Pre}

This appendix details the four visual quality metrics and pair selection procedure used to construct video-mismatch pairs for AVM-Pref.
A clip consists of a ground-truth (GT) video $\mathbf{I}=(I_0,\ldots,I_K)$ and a generated video $\hat{\mathbf{I}}=(\hat I_0,\ldots,\hat I_K)$, each with one initial frame and $K$ future frames.  
The initial frame serves as the conditioning input to the ACWM and is therefore shared, $\hat I_0=I_0$. 
The two global metrics assign one value to each future frame $t=1,\ldots,K$ and average these values over the clip. 
The two local metrics assign one value to each valid transition $t: I_{t-1}\!\rightarrow I_t$, using the optical flow between the two frames to determine the regions to score; these values are then averaged over the valid transitions of the clip.

The four metrics comprise two \emph{global} metrics that evaluate the entire frame---global PSNR $\mathrm{PSNR}^{\mathrm{glob}}$ (higher is better) and global LPIPS $\mathrm{LPIPS}^{\mathrm{glob}}$ (lower is better)---and two \emph{local} metrics that evaluate only the regions containing motion---local PSNR $\mathrm{PSNR}^{\mathrm{loc}}$ (higher is better) and local LPIPS $\mathrm{LPIPS}^{\mathrm{loc}}$ (lower is better).

\subsection{Global Visual Quality Metrics}
\label{app:four-metric-global}

Both global metrics compare each generated frame with the corresponding GT frame and average over all $K$ future frames.  
Unlike the local metrics below, they have a value on every future frame, so the average always runs over the full set $\{1,\ldots,K\}$.

\paragraph{Global PSNR.}
For frame $t$, MSE is computed over all pixels and RGB channels:
\begin{equation}
  \mathrm{MSE}_t=\frac{1}{3HW}\sum_{k=1}^{3}\sum_{i=1}^{H}\sum_{j=1}^{W}
  \bigl(I_t[k,i,j]-\hat I_t[k,i,j]\bigr)^2,
  \qquad
  \mathrm{PSNR}^{\mathrm{glob}}_t=10\log_{10}\frac{255^2}{\mathrm{MSE}_t},
  \label{eq:four-metric-global-psnr}
\end{equation}
with the convention $\mathrm{PSNR}^{\mathrm{glob}}_t=99$ whenever $\mathrm{MSE}_t\leq10^{-12}$, so that an exactly reproduced frame yields a finite score.
The global PSNR of the clip is
\begin{equation}
  \mathrm{PSNR}^{\mathrm{glob}}=\frac{1}{K}\sum_{t=1}^{K}\mathrm{PSNR}^{\mathrm{glob}}_t .
  \label{eq:four-metric-global-psnr-mean}
\end{equation}

\paragraph{Global LPIPS.}
We use the official LPIPS~\citep{zhang2018perceptual} implementation with the AlexNet backbone, loaded from a fixed local checkpoint so that every clip in the dataset is scored by identical weights.  
Frames are converted to RGB, scaled to $[0,1]$, and then affinely mapped once onto the network input range $[-1,1]$. 
Writing $d_{\mathrm{LPIPS}}$ for the distance this network produces, the global LPIPS of the
clip is
\begin{equation}
  \mathrm{LPIPS}^{\mathrm{glob}}
  =\frac{1}{K}\sum_{t=1}^{K}d_{\mathrm{LPIPS}}\bigl(\hat I_t,\,I_t\bigr).
  \label{eq:four-metric-global-lpips}
\end{equation}

\subsection{Motion Region Extraction}
\label{app:four-metric-motion}

Action-induced changes are confined to the local regions where the robot interacts with objects, and these are also the regions an ACWM is most likely to render incorrectly.
Because the two global metrics average over the whole frame, the surrounding static background dilutes this evidence, making them less sensitive to the failures that matter most.
The two local metrics therefore restrict their evaluation to these local regions.  
These regions are located from optical flow, estimated independently on the GT and on the generated video, so that the local metrics penalize both motion the ACWM fails to reproduce and motion it invents.

\paragraph{Optical Flow.}
We estimate optical flow with MegaFlow~\citep{zhang2026megaflowzeroshotlargedisplacement}.
For each video, we run the estimator on every pair of adjacent frames, obtaining fields:
\begin{equation}
  \Phi_t=\mathrm{MegaFlow}(I_{t-1},I_t)\in\mathbb{R}^{2\times H\times W},
  \qquad
  \widehat\Phi_t=\mathrm{MegaFlow}(\hat I_{t-1},\hat I_t),
  \qquad t=1,\ldots,K .
  \label{eq:four-metric-flow}
\end{equation}

\paragraph{Motion Bounding Box Extraction.}
The optical flow field $\Phi_t$ maps pixels from frame $t-1$ to frame $t$.
It is defined on the grid of frame $t-1$, where each vector $\Phi_t[:,i,j]$ gives the displacement of pixel $(i,j)$ between the two frames.
We identify moving pixels by thresholding the displacement magnitude at $\tau=2$ pixels:
\begin{equation}
  S_t=\bigl\{(i,j):\;\lVert \Phi_t[:,i,j]\rVert_2>\tau\bigr\}.
  \label{eq:four-metric-moving-pixels}
\end{equation}
Thus, $S_t$ contains the source positions of moving pixels in frame $t-1$.
Their corresponding destination positions in frame $t$ are
\begin{equation}
  S^{\rightarrow}_t=
  \Bigl\{
    \bigl(
      \mathrm{clip}(i+\lfloor \Phi^{\mathrm{v}}_t[i,j]\rceil),\,
      \mathrm{clip}(j+\lfloor \Phi^{\mathrm{u}}_t[i,j]\rceil)
    \bigr)
    :\;(i,j)\in S_t
  \Bigr\},
  \label{eq:four-metric-mask}
\end{equation}
where $\Phi^{\mathrm{u}}_t$ and $\Phi^{\mathrm{v}}_t$ denote the horizontal and vertical components of the optical flow at time $t$, respectively. 
Here, $\lfloor\cdot\rceil$ denotes rounding to the nearest integer, and $\mathrm{clip}$ restricts each coordinate to the valid range of the corresponding image axis. 
Using the same construction, we obtain the pixel sets $\widehat S_t$ and $\widehat S^{\rightarrow}_t$ for the generated video from its optical flow field $\widehat\Phi_t$.

Since the resulting pixel sets may contain isolated or spatially scattered pixels, we group neighboring pixels into rectangular regions so that each local score corresponds to a coherent moving structure. 
Specifically, for a pixel set $S$, let $\mathrm{Box}(S)$ denote the collection of bounding boxes of its connected components, with one box assigned to each component. 
We discard any box whose area is smaller than $5\times10^{-4}$ of the total frame area. 
The boxes extracted from the GT and generated videos are then merged frame by frame:
\begin{equation}
    \mathcal{B}_t=\mathrm{Box}(S_t)\cup\mathrm{Box}(\widehat S_t),
    \qquad
    \mathcal{B}^{\rightarrow}_t
    =\mathrm{Box}(S^{\rightarrow}_t)
    \cup\mathrm{Box}(\widehat S^{\rightarrow}_t),
    \label{eq:four-metric-boxes}
  \end{equation}
where $\mathcal{B}_t$ is defined on frame $t-1$ and $\mathcal{B}^{\rightarrow}_t$ on frame $t$. 
Including boxes from both videos captures complementary errors: GT boxes identify motion that the ACWM fails to reproduce, whereas generated-video boxes identify motion that it introduces spuriously.

\subsection{Local Visual Quality Metrics}
\label{app:four-metric-local}

\paragraph{Scored Regions.}
For each transition $t$, we evaluate only the bounding boxes constructed above. The boxes in $\mathcal{B}_t$ are evaluated on the source frame $t-1$, and those in $\mathcal{B}^{\rightarrow}_t$ on the destination frame $t$.
The box-level evaluations can be recorded as
\begin{equation}
  \mathcal{P}_t=
  \bigl\{(b,t-1):b\in\mathcal{B}_t\bigr\}\cup
  \bigl\{(b,t):b\in\mathcal{B}^{\rightarrow}_t\bigr\},
  \label{eq:four-metric-scored-regions}
\end{equation}
where each pair specifies one box and the frame on which it is evaluated.
For each metric, we first aggregate the box-level values within $\mathcal{B}_t$ and within $\mathcal{B}^{\rightarrow}_t$ separately, and then aggregate these two set-level values into the score of transition $t$.
For each transition, we compute local metrics over the available bounding boxes, omitting any empty box set from aggregation.
For $t=1$, the shared initial frame is excluded from local scoring, so only boxes in $\mathcal{B}^{\rightarrow}_1$ contribute.
We determine the valid transition set $\mathcal{T}_{\mathrm{valid}}$ separately for each GT-generated video comparison, skipping $t$ if neither video provides a valid scored box and discarding the clip only if this set is empty.
Thus, a transition with no valid motion region in either video is omitted, whereas spurious motion in the generated video is still evaluated even when the GT is stationary.

\paragraph{Local PSNR.}
For box $b$ in frame $f$, we compute PSNR over all enclosed pixels:
\begin{equation}
\begin{gathered}
  \mathrm{PSNR}\bigl(b,f\bigr)=10\log_{10}\frac{255^2}
  {\dfrac{1}{3\lvert b\rvert}\displaystyle\sum_{k=1}^{3}
   \sum_{(i,j)\in b}\bigl(I_f[k,i,j]-\hat I_f[k,i,j]\bigr)^2},
  \\
  \begin{aligned}
  &\mathrm{PSNR}^{\mathrm{src}}_t
  =\min_{b\in\mathcal{B}_t}\mathrm{PSNR}(b,t-1), \\
  &\mathrm{PSNR}^{\rightarrow}_t
  =\min_{b\in\mathcal{B}^{\rightarrow}_t}\mathrm{PSNR}(b,t), \\
  &\mathrm{PSNR}^{\mathrm{loc}}_t
  =\min\bigl(\mathrm{PSNR}^{\mathrm{src}}_t,\mathrm{PSNR}^{\rightarrow}_t\bigr),
  \end{aligned}
\end{gathered}
  \label{eq:four-metric-local-psnr}
\end{equation}
As with global PSNR, we set the box-level PSNR to 99 whenever the corresponding MSE is at most $10^{-12}$.
The local PSNR of the clip is
\begin{equation}
  \mathrm{PSNR}^{\mathrm{loc}}
  =\frac{1}{\lvert\mathcal{T}_{\mathrm{valid}}\rvert}\sum_{t\in\mathcal{T}_{\mathrm{valid}}}\mathrm{PSNR}^{\mathrm{loc}}_t.
  \label{eq:four-metric-local-psnr-mean}
\end{equation}

\paragraph{Local LPIPS.}
For each box $b$ evaluated on frame $f$, we crop the corresponding regions from both videos and, where necessary, pad each spatial dimension to at least 31 pixels, the minimum required by the AlexNet feature extractor used in LPIPS.
Writing $I_f\vert_b$ for the padded crop extracted from frame $f$ using box $b$,
\begin{equation}
\begin{aligned}
  &\mathrm{LPIPS}\bigl(b,f\bigr)
  =d_{\mathrm{LPIPS}}\bigl(\hat I_f\vert_b,\,I_f\vert_b\bigr), \\
  &\mathrm{LPIPS}^{\mathrm{src}}_t
  =\max_{b\in\mathcal{B}_t}\mathrm{LPIPS}(b,t-1), \\
  &\mathrm{LPIPS}^{\rightarrow}_t
  =\max_{b\in\mathcal{B}^{\rightarrow}_t}\mathrm{LPIPS}(b,t), \\
  &\mathrm{LPIPS}^{\mathrm{loc}}_t
  =\max\bigl(\mathrm{LPIPS}^{\mathrm{src}}_t,\mathrm{LPIPS}^{\rightarrow}_t\bigr).
\end{aligned}
  \label{eq:four-metric-local-lpips}
\end{equation}
The local LPIPS of the clip is
\begin{equation}
  \mathrm{LPIPS}^{\mathrm{loc}}
  =\frac{1}{\lvert\mathcal{T}_{\mathrm{valid}}\rvert}\sum_{t\in\mathcal{T}_{\mathrm{valid}}}\mathrm{LPIPS}^{\mathrm{loc}}_t.
  \label{eq:four-metric-local-lpips-mean}
\end{equation}

\subsection{Pair Selection}
\label{app:four-metric-pairing}

A pair is retained only if the positive video strictly beats the negative one on all four metrics: higher global and local PSNR and lower global and local LPIPS.
Pairs that do not satisfy all four inequalities are discarded.

We aim for comparable pair counts across models and tasks for intra-model pairs, and across model combinations and tasks for cross-model pairs; the resulting distributions are reported in Appendix~\ref{app:statistics}, as shown in Figure~\ref{fig:video_mismatch_pref}.

\section{Reward Model Implementation Details}
\label{app:reward-model-implementation}
For dynamics pretraining, preference training, and IDM training, we hold out internal validation subsets from $\mathrm{Agi_A^{train}}$ for checkpoint selection and exclude them from parameter optimization.
These internal validation subsets are distinct from $\mathrm{Agi_A^{test}}$, which is used for AVM-Eval evaluation and does not participate in checkpoint selection.

\subsection{Dynamics Pretraining}
\label{app:predictor}
This appendix specifies the input processing, predictor architecture, and dynamics-pretraining setup.
The input processing and predictor architecture are shared by both training stages.

\paragraph{Input Representation.}
We use 13-frame clips sampled at 8 fps, with one initial frame and $K=12$ future frames.
Let $\mathbf{I}=(I_0,\ldots,I_K)$ denote a clip and $\mathbf{a}=(a_1,\ldots,a_K)$ its associated action chunk, which consists of relative end-effector actions for both arms.
The action $a_t$ corresponds to the transition from frame $I_{t-1}$ to frame $I_t$.
Let $\mathbf{s}=(s_0,\ldots,s_{K-1})$ denote the corresponding state sequence, where $s_t$ records the end-effector states of both arms at frame $I_t$.
For each arm $r\in\{L,R\}$, $a_t^r\in\mathbb{R}^{7}$ comprises a 3D translation increment, a 3D rotation-vector increment, and a gripper-opening increment.
The absolute end-effector state $s_t^r\in\mathbb{R}^{10}$ comprises a 3D position, a continuous 6D rotation representation, and the gripper state.
For predictor input, we group components by arm and write $a_t=(a_t^L,a_t^R)$ and $s_t=(s_t^L,s_t^R)$.
This arm-wise action layout differs from Eq.~\ref{eq:local-action-layout}, which groups the two arms' translations, rotations, and gripper increments by component type.
The two action layouts are related by a fixed permutation of components; this permutation changes their arrangement, not their values.
The coordinate transformations and normalization used for predictor input are described below.

We first transform the absolute end-effector position and orientation components of each arm's state into the head-camera coordinate frame.
For each arm, let $p_t$ and $Q_t\in\mathrm{SO}(3)$ denote the transformed position and orientation.
The state stores $p_t$ and the first two columns of $Q_t$ as its position and 6D orientation representation, respectively.
We then compute the translation and rotation components of the action from consecutive transformed states.
Translation increments are $\Delta p_t=p_t-p_{t-1}$.
Rotation increments use left composition, $\Delta Q_t=Q_tQ_{t-1}^{\top}$, and are encoded as rotation vectors $\omega_t$ in radians.
For the gripper, we normalize the opening $w_t$ as $g_t=\operatorname{clip}(w_t/100,0,1)$.
The state stores the absolute normalized opening $g_t$, while the action stores its increment $\Delta g_t=g_t-g_{t-1}$.

We then normalize the robot inputs by scaling the translation components of both arms' actions by $65.48$ and the rotation-vector components by $11.80$, leaving the gripper command unchanged.
These coefficients are the reciprocals of the median per-step translation magnitude ($0.015271$ m) and rotation magnitude ($0.084724$ rad) computed from the training trajectories.
This scaling brings the median magnitudes of both types of action increments to approximately one.
We apply the same translation scaling factor of $65.48$ to the XYZ components of the absolute states, since state positions and action translation increments are expressed in the same head-camera coordinate frame and share the same physical units (meters).
The states' 6D orientation representations consist of two orthonormal columns of a rotation matrix, whose dimensionless entries lie in $[-1,1]$.
These representations already have a fixed scale, so we retain them without additional scaling.
Gripper states are already normalized to $[0,1]$, and their increments lie in $[-1,1]$, so neither needs rescaling.

Each full frame is resized to $384\times384$ using bilinear interpolation, without cropping.
We encode frames independently with the frozen V-JEPA 2.1 ViT-G/16 encoder, which has 48 layers, hidden dimension $D=1664$, and approximately 1.84B parameters.
With a patch size of $16\times16$, each frame produces $N=24\times24=576$ tokens of dimension $D$.
Layer normalization gives the per-frame target features
\begin{equation}
    z_t=\operatorname{LN}(E(I_t))\in\mathbb{R}^{N\times D}.
    \label{eq:acwm-frame-feature}
\end{equation}
To form the predictor input tokens, we map the normalized states, normalized actions, and visual features to the predictor's hidden dimension $d_P=1024$ using separate affine projections $f_s$, $f_a$, and $f_z$, respectively.
Let $\bar s_t^r\in\mathbb{R}^{10}$ and $\bar a_{t+1}^r\in\mathbb{R}^{7}$ denote the normalized state and action inputs for arm $r$.
We then add learnable arm-identity embeddings $e_L,e_R\in\mathbb{R}^{d_P}$ to the corresponding projected state and action features, yielding
\begin{equation}
    u^s_{t,r}=f_s(\bar s_t^r)+e_r,\qquad
    u^a_{t,r}=f_a(\bar a_{t+1}^r)+e_r.
\end{equation}
Here, $u^s_{t,r},u^a_{t,r}\in\mathbb{R}^{d_P}$ denote embedded tokens.
The same identity embedding is added to both token types for a given arm.
The visual token block is
\begin{equation}
    U^z_t=f_z(z_t)\in\mathbb{R}^{N\times d_P},
\end{equation}
where $f_z$ is applied independently to each visual patch.
Within each time block, we place the left-arm state and action tokens before the right-arm state and action tokens, followed by the visual tokens, giving
\begin{equation}
    \mathbf{X}_t=\big[u^s_{t,L},u^a_{t,L},u^s_{t,R},u^a_{t,R},U^z_t\big].
\end{equation}
Concatenating these blocks in temporal order gives the full predictor input
\begin{equation}
    \mathbf{X}=\big[\mathbf{X}_0,\ \ldots,\ \mathbf{X}_{K-1}\big].
\end{equation}
Each time block contains two state tokens, two action tokens, and $N$ visual tokens.

\paragraph{Predictor Architecture.}
The predictor has 24 Transformer layers, hidden dimension $d_P=1024$, 16 attention heads per layer, and approximately 0.31B parameters.
Block-causal attention lets a time block access itself and earlier blocks, but no future-frame blocks.
Visual patches can therefore attend to both arms' controls and the preceding visual history.
After the Transformer blocks, conditioning tokens are discarded and predicted visual tokens are projected to dimension $D=1664$ for comparison with the encoder features.

\paragraph{Training Objectives.}
We combine teacher-forcing and rollout losses, following the training scheme of V-JEPA 2-AC.
Under teacher forcing, the input blocks $\mathbf{X}_0,\ldots,\mathbf{X}_{t-1}$ contain encoded features of the observed frames rather than previous model predictions.
From these blocks, $P_\phi$ produces a prediction for frame $I_t$, which is layer-normalized to obtain $\widehat z_t^{\mathrm{TF}}\in\mathbb{R}^{N\times D}$.
The corresponding encoder feature $z_t$ serves as the supervision target.
For the rollout loss, we first predict $\widehat z_1^{\mathrm{roll}}$ from $\mathbf{X}_0$.
We then predict $\widehat z_2^{\mathrm{roll}}$ using $\mathbf{X}_0$ and $\mathbf{X}_1$, replacing only $U_1^z$ in $\mathbf{X}_1$ with $f_z(\widehat z_1^{\mathrm{roll}})$.
Both rollout predictions are layer-normalized, and gradients pass through the first prediction.
The equally weighted losses compare these predictions with the encoder targets $z_t$,
\begin{equation}
    \mathcal{L}_{\mathrm{dyn}}(\phi)
    =\underbrace{\frac{1}{KND}\sum_{t=1}^{K}
    \left\|\widehat z_t^{\mathrm{TF}}-z_t\right\|_1}_{\mathcal{L}_{\mathrm{TF}}}
    +\underbrace{\frac{1}{2ND}\sum_{t=1}^{2}
    \left\|\widehat z_t^{\mathrm{roll}}-z_t\right\|_1}_{\mathcal{L}_{\mathrm{roll}}}.
\end{equation}
Only the predictor is updated, and its learned weights initialize $P_{\phi}$ for preference training.

\paragraph{Training Setup.}
We train for 120 epochs, sampling one clip per episode in each epoch, with a global batch of 256 clips (16 per GPU on 16 GPUs).
Based on the reported 34k episodes, this corresponds to approximately 4.08M clip draws and 15.94k optimizer steps.
The learning rate warms up from $7.5\times10^{-5}$ to $3\times10^{-4}$ over the first $10\%$ of training steps, remains constant for the next $45\%$, and decays linearly to $10^{-6}$ over the final $45\%$.
Both training stages use AdamW with $(\beta_1,\beta_2)=(0.9,0.999)$, $\epsilon=10^{-8}$, weight decay $0.04$ (excluding biases and one-dimensional parameters), gradient-norm clipping at $1.0$, and bfloat16 mixed precision.
After completing the prescribed pretraining budget, we select the checkpoint with the lowest loss $\mathcal{L}_{\mathrm{dyn}}$ on the internal validation subset of $\mathrm{Agi_A^{train}}$ to initialize preference training.

\subsection{Preference Training}
\label{app:preference-training}
We initialize $P_{\phi}$ from dynamics pretraining and randomly initialize $h_{\psi}$, keeping $E$ frozen.
The clip length, input representation, and predictor architecture follow Appendix~\ref{app:predictor}.
We expand on Sec.~\ref{reward-model-architectural} by detailing the conditioning states, score head architecture, preference objective, and training setup.

\paragraph{Conditioning States.}
Following the main-text notation, we represent each candidate as $q=(\mathbf{I}',\mathbf{a}',\mathbf{s}')$, where $\mathbf{I}'$ is a real or generated video, $\mathbf{a}'=(a_1',\ldots,a_K')$ is the original or perturbed action chunk, and $\mathbf{s}'=(s_0',\ldots,s_{K-1}')$ is the corresponding state sequence.
For both original and perturbed actions, we construct $\mathbf{s}'$ by integrating the candidate's own actions from the shared initial state $s_0$:
\begin{equation}
    s_0'=s_0,\qquad s_t'=F(s_{t-1}',a_t'),\quad t=1,\ldots,K-1,
\end{equation}
where $F$ updates the end-effector state using the relative action under the input conventions in Appendix~\ref{app:predictor}.
For action-mismatch pairs, the two candidates share the same real video and initial state, but each uses its own original or perturbed action chunk to construct its conditioning states.
This keeps the states consistent with the actions and avoids conditioning the negative candidate on states obtained from the positive actions.
For video-mismatch pairs, the candidates instead contain different generated videos but share the original action chunk and initial state, so they also share the integrated conditioning states.

We use the same state construction during Flow-GRPO on $\mathrm{Agi_B}$, taking the initial state and action chunk from the AgiBotWorld recordings.
The world model is conditioned on $(I_0,\mathbf{a})$, and each generated video is scored with $\mathbf{I}'=\hat{\mathbf{I}}$, $\mathbf{a}'=\mathbf{a}$, and the states $\mathbf{s}'$ integrated from $s_0$ and $\mathbf{a}'$.
All candidates sampled under the same condition therefore share the actions and conditioning states.
ACWM-RM combines these inputs with the candidate's own visual history and evaluated frames, without requiring ground-truth future frames.

In our offline experiments, actions are derived from differences between recorded end-effector states, so integrating them from $s_0$ recovers the recorded trajectory.
The conditioning states therefore contain trajectory information already encoded in the actions and initial state.
Here, \emph{reference-free} means that scoring requires no ground-truth future video or separately supplied future measured states.

\paragraph{Score Head Architecture.}
Here we specify the dimensions along the path from a pair of patch features to the reward score.
The shared projection $W_{\mathrm{proj}}\in\mathbb{R}^{d\times D}$ maps both branches from $D=1664$ down to $d=256$ and is bias-free, so the five terms of Equation~\ref{eq:acwm-comparison-feature}, four of dimension $d$ and one scalar cosine similarity, concatenate into a comparison vector of dimension $4d+1=1025$.
Both MLPs of Equation~\ref{eq:acwm-score-and-weight-mlps} read this vector and share the architecture $1025\rightarrow256\rightarrow1$ with a GELU activation between the two affine layers, differing only in whether their output serves as a patch score or a patch weight.
Together, $W_{\mathrm{proj}}$ and the two MLPs form $h_{\psi}$ with 951,810 randomly initialized parameters.
ACWM-RM has approximately 2.15B parameters in total, including the 1.84B frozen encoder, the 0.31B predictor, and the score head.
During preference training, only the pretrained predictor and the randomly initialized score head are updated, giving approximately 0.31B trainable parameters.
Gradients reach $W_{\mathrm{proj}}$ through both branches, while the encoded features are detached so that no gradient enters $E$.
We use the unbounded scalar output of Equation~\ref{eq:acwm-clip-score} directly as the reward, without a final sigmoid.
The Bradley-Terry loss in Equation~\ref{eq:acwm-bt-loss} applies the sigmoid to the score difference $R^{+}-R^{-}$ rather than to either candidate score individually.

\paragraph{Preference Objective.}
Let $q^+$ and $q^-$ denote the preferred and dispreferred candidates, respectively, with each candidate containing the video, actions, and conditioning states defined above.
We score both candidates with shared parameters and jointly optimize $P_{\phi}$ and $h_{\psi}$ using
\begin{equation}
    \mathcal{L}_{\mathrm{BT}}(\phi,\psi)
    =-\mathbb{E}_{(q^+,q^-)\sim\mathrm{AVM\text{-}Pref}}
    \left[\log\sigma(R^+-R^-)\right],
    \label{eq:acwm-bt-loss}
\end{equation}
where $\sigma$ is the sigmoid function and $R^{\pm}=R_{\phi,\psi}(q^{\pm})$.
For numerical stability, the per-pair loss is implemented as $\operatorname{softplus}(-(R^+-R^-))$.

\paragraph{Training Setup.}
We use the shared optimizer settings in Appendix~\ref{app:predictor}, with a global batch of 256 preference pairs (16 per GPU on 16 GPUs).
Training runs for 3 epochs, corresponding to approximately 1.52M preference-pair draws, 3.04M candidate-clip draws, and 5.93k optimizer steps based on the reported 506.3k pairs.
The head learning rate warms up from $10^{-5}$ to $4\times10^{-4}$ over the first $5\%$ of training steps, remains constant for the next $20\%$, and decays to $4\times10^{-5}$ with a cosine schedule over the final $75\%$.
The predictor learning rate is $0.05$ times the head rate throughout training, including warmup.
After completing the prescribed preference-training budget, we select the checkpoint with the highest pairwise accuracy on internal validation pairs from $\mathrm{Agi_A^{train}}$.
\subsection{Training Budget Accounting}
\label{app:training-budget}
We report nominal training budgets in 13-frame clip draws, including repeated samples, based on corpus sizes before internal validation holdout.
Each pretraining clip contributes one draw, while each preference pair contributes two candidate-clip draws.
For each IDM, one draw likewise corresponds to a complete clip covering all 12 action transitions; predictions within a clip are not counted as separate draws.
Pretraining samples one clip per episode per epoch, so 120 epochs over 34k episodes yield $120\times34{,}000=4.08\mathrm{M}$ clip draws.
Preference training traverses 506.3k pairs for 3 epochs, yielding $3\times506{,}300=1{,}518{,}900$ pair draws, or $3{,}037{,}800\approx3.04\mathrm{M}$ candidate-clip draws.
Together, the two stages have a nominal budget of
\[
    120\times34{,}000+3\times506{,}300\times2=7{,}117{,}800\approx7.12\mathrm{M}
\]
training clip draws, approximately matching the nominal budget of each IDM in Table~\ref{tab:idm-optim}.
Actual draw and step counts depend on the held-out subsets and final partial batches.
This accounting matches nominal clip-draw budgets and temporal coverage, not computational cost or the number of unique videos or frames seen by each model.

\section{Training Setup for the Baseline IDMs}
\label{app:idm-training-setups}
We train four baseline IDMs on $\mathrm{Agi_A^{train}}$, starting from the pretrained backbone checkpoints used in their official implementations.
Specifically, EVA-IDM uses pretrained ResNet-50, GR00T-IDM uses pretrained SigLIP2-L/16, TAP-IDM uses pretrained Qwen2.5-VL-3B, and J-IDM uses pretrained VGGT.
For each baseline, we follow its official implementation with the bimanual action-interface adaptations described below; the resulting architectures, observation-action interfaces, and hyperparameters are summarized in Tables~\ref{tab:idm-arch} and~\ref{tab:idm-optim}.
TAP-IDM retains its original LM head and shared action-token vocabulary, configuring the FAST tokenizer to jointly decode 14-dimensional bimanual actions.
J-IDM uses a 14-dimensional command space and a 28-channel per-pixel Jacobian head, jointly recovering both arms' actions through ridge-regularized inversion.
GR00T-IDM retains its 32-dimensional output with 20 valid components and padding masked during training; before reward computation, padding is discarded, and each arm's 6D rotation is converted via quaternions to a 3D rotation-vector increment, yielding the 14-dimensional action layout in Eq.~\ref{eq:local-action-layout}.
Its action horizon is set to 12, matching the 12 transitions in each 13-frame clip.

For each IDM, we set a nominal budget of approximately 7.12M training clip draws, comparable to ACWM-RM's combined budget (Appendix~\ref{app:training-budget}).
Each sampled clip contains 13 frames and 12 action transitions. We apply each IDM's native observation window across the clip, with all 12 actions contributing to its training loss.
After completing this budget, we select each IDM's checkpoint by the lowest mean distance between its predicted actions and the ground-truth actions on the internal validation subset of $\mathrm{Agi_A^{train}}$.

\begin{table}[!t]
  \centering
  \footnotesize
  \setlength{\tabcolsep}{4pt}
  \caption{Architectures and observation-action interfaces. Frame counts are per prediction within a 13-frame training clip.}
  \label{tab:idm-arch}
  \begin{tabular}{@{}lcccc@{}}
    \toprule
     & EVA-IDM & GR00T-IDM & TAP-IDM & J-IDM \\
    \midrule
    Visual backbone
      & ResNet-50 & SigLIP2-L/16 & Qwen2.5-VL-3B & VGGT \\
    Backbone initialization
      & pretrained & pretrained & pretrained & pretrained \\
    Input resolution
      & $512\times512$ & $224\times224$ & $224\times224$ & $128\times128$ \\
    Frames/prediction
      & 3 & 2 & 2 & 2 \\
    Action dim for reward
      & (14) & (14) & (14) & (14) \\
    Objective
      & smooth-$L_1$ & flow matching & CE & Charbonnier flow \\
    Action recovery
      & direct & 16-step Euler & detokenize & damped least sq \\
    \bottomrule
  \end{tabular}
\end{table}

\begin{table}[!t]
  \centering
  \footnotesize
  \setlength{\tabcolsep}{5pt}
  \caption{IDM optimization settings. Each model uses 8 GPUs; batch sizes count complete 13-frame clips. Optimizer steps are set to target a nominal budget
  of approximately 7.12M clip draws per model. Step and draw counts are rounded.}
  \label{tab:idm-optim}
  \begin{tabular}{@{}lcccc@{}}
    \toprule
     & EVA-IDM & GR00T-IDM & TAP-IDM & J-IDM \\
    \midrule
    Batch / GPU
      & 32 & 16 & 16 & 16 \\
    GPUs
      & 8 & 8 & 8 & 8 \\
    Global batch
      & 256 & 128 & 128 & 128 \\
    Learning rate
      & $3.0\!\times\!10^{-4}$ & $1.0\!\times\!10^{-4}$ & $5.0\!\times\!10^{-5}$ & $5.0\!\times\!10^{-5}$ \\
    Steps
      & 27.8k & 55.6k & 55.6k & 55.6k \\
    Clip draws
      & 7.12M & 7.12M & 7.12M & 7.12M \\
    Optimizer
      & AdamW & AdamW & AdamW & AdamW \\
    Weight decay
      & -- & $1\!\times\!10^{-5}$ & $0.05$ & $1\!\times\!10^{-5}$ \\
    Betas
      & -- & $(0.95,0.999)$ & $(0.9,0.999)$ & $(0.9,0.99)$ \\
    Schedule
      & cosine & cosine & cosine & constant \\
    Warmup
      & $5\%$ & $5\%$ & 1k steps & -- \\
    Precision
      & bf16 & bf16 & bf16 & fp16 \\
    Gradient clip
      & -- & -- & -- & $1.0$ \\
    \bottomrule
  \end{tabular}
\end{table}

\paragraph{Action Standardization and IDM Reward Computation.}
\label{app:idm-reward-normalization}
For the IDM-based Flow-GRPO baseline, we standardize both the predicted and conditioning actions using per-dimension statistics estimated from action samples drawn from $\mathrm{Agi_A^{train}}$.
We compute the empirical mean $\mu_j$ and standard deviation $\sigma_j$ separately for each of the 14 camera-frame delta-action dimensions and keep these statistics fixed during reward computation.
Table~\ref{tab:idm-action-std} reports the standard deviations used in the IDM reward.
Following the notation in Section~3, $\tilde a_{k,j}$ and $a_{k,j}$ denote the predicted and conditioning actions at transition $k\in\{1,\ldots,K\}$ and dimension $j\in\{1,\ldots,14\}$.
Their standardized values and residuals are
\begin{equation}
\begin{aligned}
    \bar{\tilde a}_{k,j}&=\frac{\tilde a_{k,j}-\mu_j}{\sigma_j+10^{-8}},
    &\bar a_{k,j}&=\frac{a_{k,j}-\mu_j}{\sigma_j+10^{-8}},\\
    e_{k,j}&=\bar{\tilde a}_{k,j}-\bar a_{k,j}
    =\frac{\tilde a_{k,j}-a_{k,j}}{\sigma_j+10^{-8}}.
\end{aligned}
    \label{eq:idm-action-normalization}
\end{equation}
The same statistics are applied to both actions, so the mean cancels in the residual.

\begin{table}[!t]
  \centering
  \footnotesize
  \setlength{\tabcolsep}{5pt}
  \caption{Per-dimension action standard deviations used for IDM reward computation, estimated from AgiBotWorld.}
  \label{tab:idm-action-std}
  \begin{tabular*}{0.66\linewidth}{@{\extracolsep{\fill}}lccc@{}}
    \toprule
    Action component & Unit & Left arm $\sigma_j$ & Right arm $\sigma_j$ \\
    \midrule
    X translation & m & 0.003746 & 0.004671 \\
    Y translation & m & 0.003482 & 0.003662 \\
    Z translation & m & 0.003115 & 0.003379 \\
    X rotation & rad & 0.014987 & 0.013658 \\
    Y rotation & rad & 0.013690 & 0.014198 \\
    Z rotation & rad & 0.015895 & 0.015860 \\
    Gripper-opening change & -- & 0.078766 & 0.084997 \\
    \bottomrule
  \end{tabular*}
\end{table}

We then apply the elementwise Smooth-L1 loss with threshold 1 and average over all transitions and action dimensions:
\begin{equation}
\begin{gathered}
    \ell_{\mathrm{SL1}}(e)=
    \begin{cases}
        \tfrac12 e^2, & |e|<1,\\
        |e|-\tfrac12, & |e|\geq1,
    \end{cases}\\
    R^{\mathrm{IDM}}
    =-\frac{1}{14K}\sum_{k=1}^{K}\sum_{j=1}^{14}
    \ell_{\mathrm{SL1}}(e_{k,j}).
\end{gathered}
    \label{eq:idm-normalized-reward}
\end{equation}
Here, $K$ is the number of action transitions, with $K=12$ for the 13-frame clips used in our experiments.
We average the elementwise Smooth-L1 losses over all (K) transitions and 14 action dimensions with equal weights.
The reward scale is 1.0, yielding one nonpositive scalar per video; values closer to zero indicate better action agreement.

Dividing by the per-dimension standard deviation expresses translation, rotation, and gripper errors on a common dimensionless scale relative to their variability in $\mathrm{Agi_A^{train}}$.
Smooth-L1 penalizes small standardized residuals quadratically and large residuals linearly, reducing the influence of large errors relative to a squared-error loss.

\section{Prompt and Criteria for Evaluation}
\label{app:prompts_and_criteria}

\begin{table*}[!t]
\centering
\resizebox{0.97\textwidth}{!}{%
\begin{tcolorbox}[width=\textwidth]
You are a strict evaluator for a short generated robot-manipulation video. The input is a 13-frame clip; the first frame is the conditioning observation and the remaining 12 frames are generated.
Given the task description, inspect every frame and score only the visible evidence.
Evaluate: (i) \emph{object consistency}, checking whether the manipulated object preserves its identity, shape, structure, color, material, and count;
(ii) \emph{manipulator consistency}, checking whether the robot, arms, hands, and grippers preserve their identity, geometry, connectivity, number of parts, and plausible articulation; 
and (iii) \emph{anomaly check}, detecting floating or penetrating entities, teleportation, spontaneous appearance or disappearance, duplication, unsupported attachment, and false grasps. 
Assign each requested dimension a score from 1 (severe failure) to 5 (fully plausible and stable), using only information visible in the clip. Do not evaluate task completion or the final goal. Return a JSON object containing the scores and a brief evidence-based reason for each dimension.
\end{tcolorbox}}
\caption{Prompt used by Qwen3-VL to compute rewards during Flow-GRPO and by GPT-5.6 sol to evaluate the physical consistency of videos
  generated by DreamDojo after Flow-GRPO fine-tuning.}
\label{tab:vlm-prompt}
\end{table*}

When evaluating ACWM-generated videos for visual artifacts and physical inconsistencies, or using the VLM-based rewards to train the ACWM with Flow-GRPO, we use the evaluation prompt from RBench~\citep{deng2026rethinking} shown in Table~\ref{tab:vlm-prompt}.
The prompt produces three criteria: Object Consistency (OC), Manipulator Consistency (MC), and Anomaly Check (AC), each rated on an integer scale from 1 to 5.

For the VLM-based rewards, we normalize each score to $[0,1]$ and average the three normalized scores with equal weights to obtain the scalar reward used by Flow-GRPO:
\[
 S_{\mathrm{VLM}}=\frac{1}{3}\left(
 \frac{\mathrm{OC}-1}{4}
 +\frac{\mathrm{MC}-1}{4}
 +\frac{\mathrm{AC}-1}{4}
 \right).
\]

The three criteria are defined as follows:
\begin{itemize}[leftmargin=*]
\item \emph{Object Consistency} measures whether the target object preserves its identity and attributes throughout the video, including its shape, appearance, material, and count.
\item \emph{Manipulator Consistency} measures the temporal stability of the robot embodiment, including its morphology, link or manipulator connectivity, joint articulation, and structural stability.
\item \emph{Anomaly Check} measures physical and commonsense violations, such as floating or unsupported entities, interpenetration, spontaneous appearance or disappearance, duplication, non-contact attachment, and incorrect grasping.
\end{itemize}

\section{Flow-GRPO Optimization Details}
\label{app:flow_grpo_details}
We fine-tune the policy using LoRA with rank 32 on 16 NVIDIA H800 GPUs.
We optimize the LoRA parameters with AdamW at a learning rate of $3\times10^{-4}$ for 120 outer training iterations.
Each rollout uses 20 Flow-GRPO denoising steps and a diffusion noise level of $0.7$. 
Each sampled episode from $\mathrm{Agi_B}$ (Table~\ref{tab:tasks_ood_agib}) supplies a shared input condition---the initial frame $I_0$ and action chunk $\mathbf{a}$---for a group of $G=8$ rollouts.
In each outer iteration, we collect four rollout batches with a per-GPU sampling batch size of 8 episodes and group size $G$, yielding $8\times8=64$ rollouts per batch and 256 rollouts per GPU per outer iteration.
We then perform four inner PPO-style policy-optimization epochs on the collected data and accumulate gradients over 16 rollouts.
The PPO-style ratio clipping range is $10^{-3}$, the KL penalty coefficient is $\beta=0.004$, and gradients are clipped to a maximum norm of 1.0. 
We use seed 42 for policy optimization.
ACWM-RM is frozen throughout policy optimization and uses the conditioning state sequence integrated from the initial state and each sampled action chunk.

For evaluation on $\mathrm{Agi_C}$, we generate videos from each optimized ACWM using five video-generation seeds and report the mean and standard deviation across seeds for visual quality, action-following fidelity, and physical consistency.

\section{Dataset and Preference Pair Statistics}
\label{app:statistics}
\raggedbottom
\setlength{\intextsep}{4pt}
\setlength{\abovecaptionskip}{4pt}
\setlength{\belowcaptionskip}{3pt}

\paragraph{Source datasets.}
Tables~\ref{tab:tasks_id}, \ref{tab:tasks_ood_agib}, and~\ref{tab:tasks_agic} list the tasks and episode counts in $\mathrm{Agi_A^{train}}$, $\mathrm{Agi_B}$, and $\mathrm{Agi_C}$, respectively.
Tables~\ref{tab:tasks_ood_robot}, \ref{tab:tasks_ood_gen}, and~\ref{tab:tasks_ood_robod} report the corresponding statistics for $\mathrm{Robo_T}$, $\mathrm{Gen}$, and $\mathrm{Robo_D}$ used in AVM-Eval.

\begin{figure}[H]
\centering
\vspace{-6pt}
\includegraphics[width=0.90\textwidth]{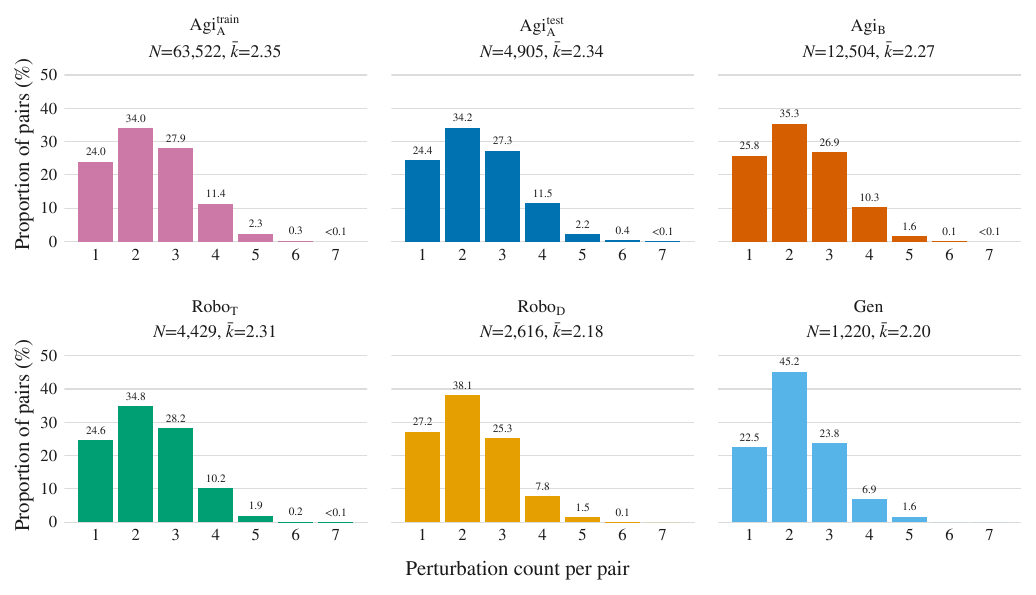}
\caption{
Distribution of numerical perturbation counts per action-mismatch pair.
$\mathrm{Agi_A^{train}}$ shows AVM-Pref; all other panels show AVM-Eval. 
$N$ is the number of distinct numerical-perturbation pairs, and $\bar{k}$ is the mean number of distinct perturbation types per pair.}
\label{fig:pert_arity}
\vspace{-6pt}
\end{figure}

\paragraph{Action-mismatch pairs.}
Tables~\ref{tab:pert_numeric} and~\ref{tab:pert_semantic} summarize pair counts in AVM-Eval and AVM-Pref by numerical perturbation type and semantic perturbation group, respectively.
Figure~\ref{fig:pert_arity} shows the distribution of the number of numerical perturbation types per pair, together with the number of distinct pairs $N$ and the mean number of types per pair $\bar{k}$.

\begin{figure}[H]
\centering
\includegraphics[width=0.99\textwidth]{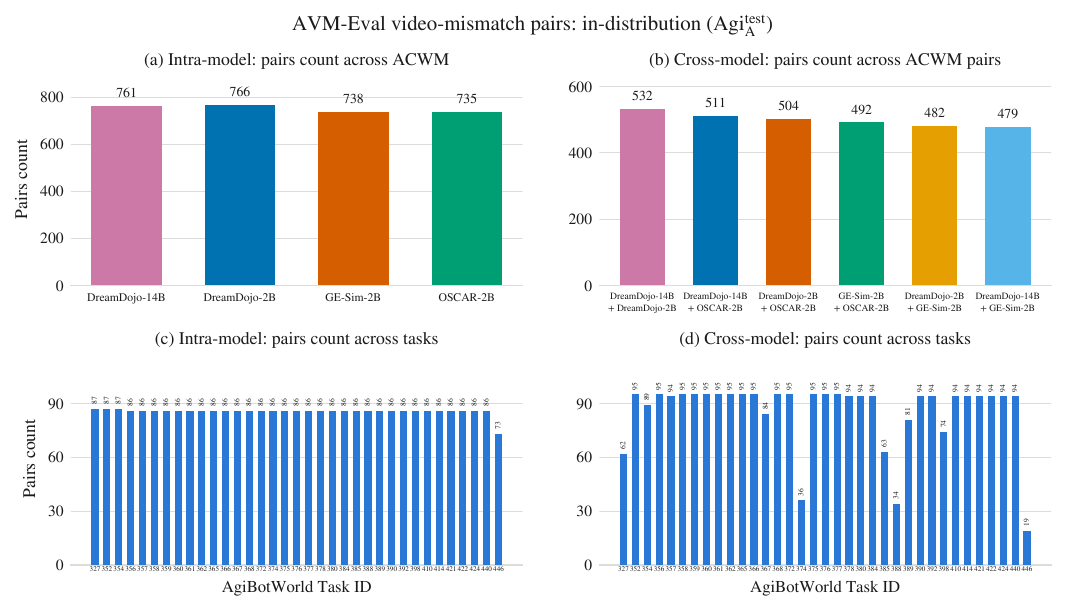}
\caption{Video-mismatch pair statistics for AVM-Eval on $\mathrm{Agi_A^{test}}$, with 3{,}000 intra-model and 3{,}000 cross-model pairs. Panels (a) and (b) show counts by ACWM and ACWM pair, respectively; (c) and (d) show the corresponding task distributions.}
\label{fig:video_mismatch_eval_agia}
\end{figure}

\begin{figure}[H]
\centering
\includegraphics[width=0.99\textwidth]{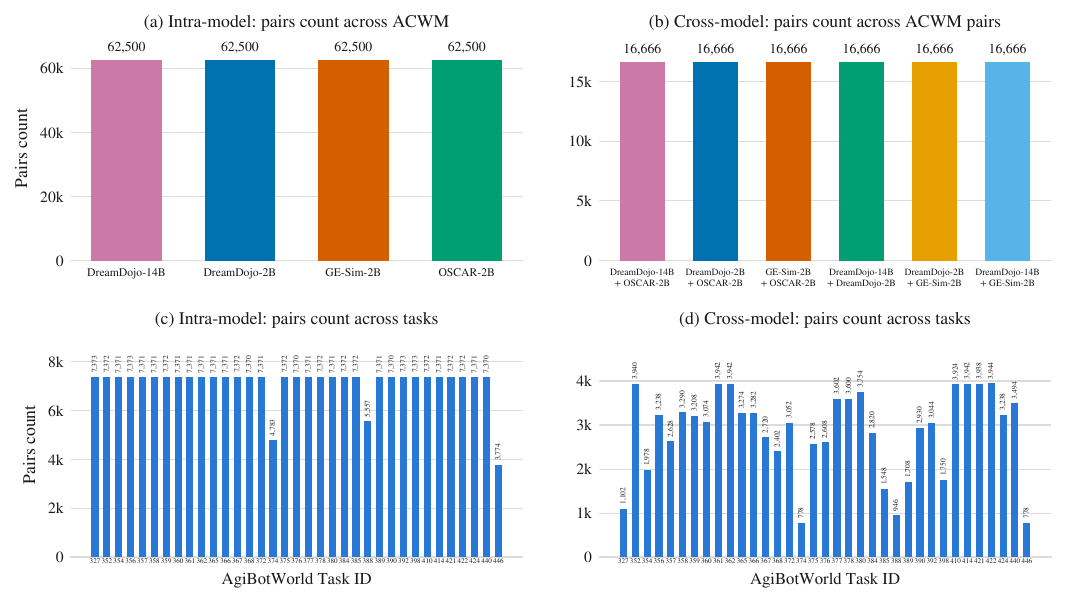}
\caption{
Statistics of video-mismatch pairs in AVM-Pref constructed from $\mathrm{Agi_A^{train}}$. 
Panels (a) and (b) show intra-model pair counts by ACWM and cross-model pair counts by ACWM combination, respectively; 
panels (c) and (d) show the corresponding counts by source task. 
We aim to balance pair counts across models and tasks. 
However, task-level balance is limited by the availability of candidate pairs that satisfy all four quality criteria. 
Cross-model pairing additionally requires rollouts from both ACWMs under the same input condition, further restricting the available candidates for some tasks. 
So, tasks with insufficient eligible pairs cannot meet their target quotas, and the remaining quotas are redistributed to tasks with sufficient candidates.
}
\label{fig:video_mismatch_pref}
\end{figure}

\begin{figure}[H]
\centering
\includegraphics[width=0.92\textwidth]{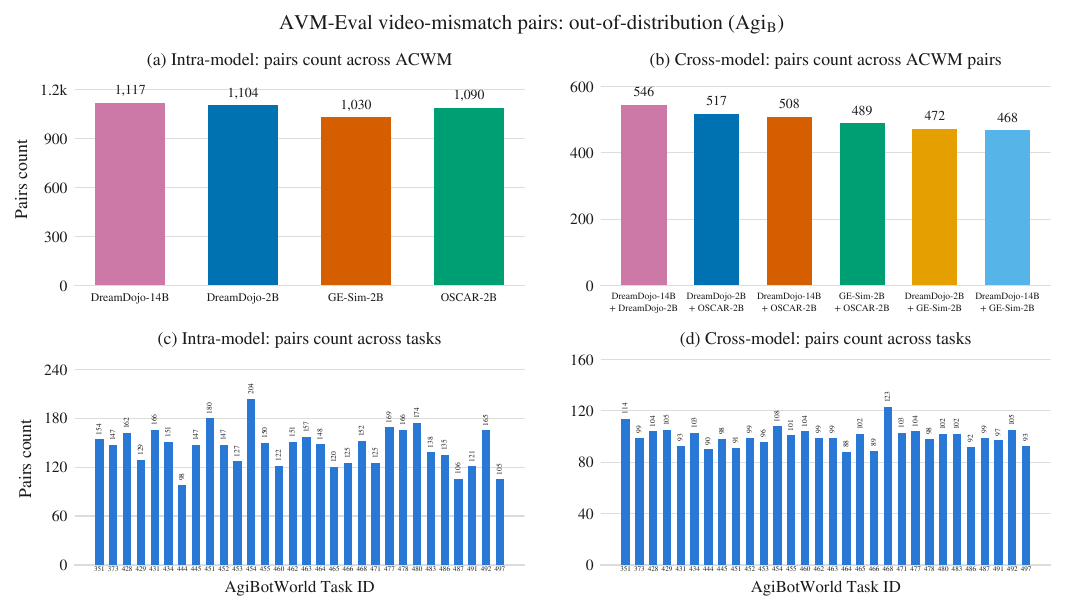}
\caption{Statistics of video-mismatch pairs in AVM-Eval on $\mathrm{Agi_B}$, comprising 4{,}341 intra-model and 3{,}000 cross-model pairs. Panels (a) and (b) show counts by ACWM and ACWM pair, respectively; panels (c) and (d) show the corresponding task distributions.}
\label{fig:video_mismatch_eval_agib}
\end{figure}

\paragraph{Video-mismatch pairs.}
Figure~\ref{fig:video_mismatch_pref} reports AVM-Pref statistics on $\mathrm{Agi_A^{train}}$.
Figures~\ref{fig:video_mismatch_eval_agia} and~\ref{fig:video_mismatch_eval_agib} report AVM-Eval statistics on $\mathrm{Agi_A^{test}}$ and $\mathrm{Agi_B}$, respectively.
Each figure breaks down intra-model and cross-model pairs by generating model or model pair and by task.
Task counts in Table~\ref{tab:benchmark_composition} refer to the source datasets, whereas the task distributions below include only tasks with retained video-mismatch pairs.

\begin{table}[H]\centering
\begin{adjustbox}{width=0.92\textwidth}
\begin{minipage}{\textwidth}
\centering\fontsize{8.5}{10}\selectfont
\setlength{\tabcolsep}{3.5pt}
\caption{Task-level statistics of $\mathrm{Agi_A^{train}}$, comprising 36 tasks and 33{,}981 episodes used for IDM training and ACWM-RM predictor pretraining.}
\label{tab:tasks_id}
\begin{tabularx}{\textwidth}{@{}lYr@{\hspace{0.75em}}|@{\hspace{0.75em}}lYr@{}}\toprule
ID & Task & Episodes & ID & Task & Episodes \\
\midrule
327 & Pickup items in the supermarket & 203 & 377 & Packing in e-commerce & 790 \\
352 & Open the fridge to get food. & 1{,}380 & 378 & Clear table in the restaurant & 732 \\
354 & Pickup items in the supermarket & 507 & 380 & Packing in e-commerce & 813 \\
356 & Packing in the supermarket & 473 & 384 & Insert a book into the bookshelf & 805 \\
357 & Wash dishes with dishwasher & 425 & 385 & Pickup in the supermarket produce section & 214 \\
358 & Toast bread & 469 & 388 & Pickup in the supermarket & 89 \\
359 & Sort in the warehouse & 640 & 389 & Pickup in the supermarket & 307 \\
360 & Packing in the supermarket & 513 & 390 & Checkout and scan barcode in the supermarket & 747 \\
361 & Flatten shorts & 2{,}544 & 392 & Brush water bottle & 956 \\
362 & Fold shorts & 8{,}194 & 398 & Sort clothes & 173 \\
365 & Sort personal care products & 501 & 410 & Water Pouring in Restaurant & 1{,}083 \\
366 & Sort food & 726 & 414 & Hang clothes with hanger & 2{,}043 \\
367 & Take toast from toaster & 257 & 421 & Pick up the item to wipe away the stain. & 1{,}645 \\
368 & Cook vegetables with oven & 577 & 422 & Pack items for industrial logistics & 2{,}078 \\
372 & Packing in the supermarket & 534 & 424 & Clear the countertop waste & 910 \\
374 & Sort laundry and personal care products & 91 & 440 & Iron clothes & 1{,}082 \\
375 & Brew tea & 419 & 446 & Transport table with another robot & 72 \\
376 & Sort electronic products & 402 & 475 & Ironing Clothes & 587 \\
\midrule
\multicolumn{5}{@{}l}{\textbf{Total: 36 tasks}} & \textbf{33{,}981} \\
\bottomrule
\end{tabularx}
\end{minipage}
\end{adjustbox}
\end{table}

\begin{table}[H]\centering
\begin{adjustbox}{width=0.92\textwidth}
\begin{minipage}{\textwidth}
\centering\fontsize{8.5}{10}\selectfont
\setlength{\tabcolsep}{3.5pt}
\caption{Task-level statistics of $\mathrm{Agi_B}$, comprising 30 tasks and 10{,}000 episodes used for AVM-Eval construction and Flow-GRPO fine-tuning.}
\label{tab:tasks_ood_agib}
\begin{tabularx}{\textwidth}{@{}lYr@{\hspace{0.75em}}|@{\hspace{0.75em}}lYr@{}}\toprule
ID & Task & Episodes & ID & Task & Episodes \\
\midrule
351 & Open the wardrobe and hang the clothes. & 608 & 463 & Clean the microwave and range hood. & 454 \\
373 & Sweep the floor & 583 & 464 & Replace the toilet paper roll & 428 \\
428 & Open drawer and store items & 467 & 465 & Wash clothes in the washing machine. & 329 \\
429 & Install memory module & 24 & 466 & Separate dark and light clothes & 440 \\
431 & Brew coffee with a capsule machine & 360 & 468 & Remove clothes from the drying rack and place them in the basket. & 454 \\
434 & Remove capsules from the coffee machine. & 258 & 471 & Organize the condiments on the stove. & 294 \\
444 & Fold short sleeve & 175 & 477 & Fold towels on the table & 214 \\
445 & Open the fridge to get fruits and vegetables. & 454 & 478 & Wipe the mirror cabinet & 454 \\
451 & Make a salad & 151 & 480 & Make sandwiches with salad dressing 1213 edition & 454 \\
452 & Slice a lemon to make lemon water. & 278 & 483 & Place items from the meeting room table into the storage box. & 223 \\
453 & Make a sandwich & 83 & 486 & Remove bottled water from the carton and arrange it neatly on the table. & 195 \\
454 & Make a sandwich & 116 & 487 & Remove bottled water from the carton and arrange neatly on the table. & 260 \\
455 & Serve meals & 285 & 491 & Iron clothes & 267 \\
460 & Remove the pillowcase from the clothesline and place it in the basket. & 454 & 492 & Pack in the supermarket cashier & 454 \\
462 & Discard the trash bag into the large bin. & 330 & 497 & Restock in the supermarket & 454 \\
\midrule
\multicolumn{5}{@{}l}{\textbf{Total: 30 tasks}} & \textbf{10{,}000} \\
\bottomrule
\end{tabularx}
\end{minipage}
\end{adjustbox}
\end{table}

\begin{table}[H]\centering
\begin{adjustbox}{width=0.92\textwidth}
\begin{minipage}{\textwidth}
\centering\fontsize{8.5}{10}\selectfont
\setlength{\tabcolsep}{3.5pt}
\caption{Task-level statistics of $\mathrm{Agi_C}$, comprising 10 tasks and 1{,}499 episodes used to evaluate ACWM rollouts after Flow-GRPO fine-tuning.}
\label{tab:tasks_agic}
\begin{tabularx}{\textwidth}{@{}lYr@{}}\toprule
ID & Task & Episodes \\
\midrule
363 & Open the drawer to store clothes. & 150 \\
369 & Remove clothes from the washing machine & 150 \\
425 & Prepare oatmeal porridge & 150 \\
433 & Make instant coffee & 150 \\
438 & Place the pen into the pen holder. & 150 \\
470 & Tidy the tables at the milk tea shop. & 149 \\
474 & Arrange flowers & 150 \\
485 & Pick up the scrap paper on the desk and feed it into the shredder for destruction. & 150 \\
494 & Organize the kitchen counter seasonings & 150 \\
498 & Make milk tea & 150 \\
\midrule
\multicolumn{2}{@{}l}{\textbf{Total: 10 tasks}} & \textbf{1{,}499} \\
\bottomrule
\end{tabularx}
\end{minipage}
\end{adjustbox}
\end{table}

\begin{table}[H]\centering
\begin{adjustbox}{width=0.92\textwidth}
\begin{minipage}{\textwidth}
\centering\fontsize{8.5}{8.5}\selectfont
\setlength{\tabcolsep}{3.5pt}
\caption{Task-level statistics of $\mathrm{Robo_D}$, comprising 35 tasks and 1{,}450 real-world episodes used to construct action-mismatch pairs in AVM-Eval.}
\label{tab:tasks_ood_robod}
\begin{tabularx}{\textwidth}{@{}Yr@{\hspace{0.75em}}|@{\hspace{0.75em}}Yr@{}}\toprule
Task & Episodes & Task & Episodes \\
\midrule
arrange\us largest\us number & 47 & pack\us objects\us into\us box & 43 \\
build\us tower & 48 & play\us Xylophone & 45 \\
classify\us objects & 53 & play\us stacking\us toy & 44 \\
cover\us blocks & 53 & play\us tic\us tac\us toe & 41 \\
deposit\us coin & 42 & plug\us in\us charger & 41 \\
dlc & 43 & pour\us balls\us into\us vase & 48 \\
fasten\us screws & 31 & pour\us liquid\us into\us cup & 48 \\
fill\us egg\us holder & 49 & press\us by\us number & 41 \\
fill\us pen\us holder & 43 & push\us T & 39 \\
fold\us clothes & 42 & put\us bottles\us into\us dustbin & 46 \\
hang\us mugs & 30 & sort\us nesting\us dolls\us by\us size & 43 \\
imitate\us sorting\us sequence & 35 & stack\us blocks & 15 \\
insert\us key & 36 & stack\us bowls & 48 \\
insert\us tubes & 42 & store\us laptop\us and\us headphones & 42 \\
make\us kong & 39 & swap\us T & 14 \\
make\us toast & 41 & swap\us blocks & 45 \\
match\us and\us pick\us from\us conveyor & 44 & sweep\us blocks & 45 \\
organize\us table & 44 &  &  \\
\midrule
\multicolumn{3}{@{}l}{\textbf{Total: 35 tasks}} & \textbf{1{,}450} \\
\bottomrule
\end{tabularx}
\end{minipage}
\end{adjustbox}
\end{table}

\begin{table}[H]\centering
\begin{adjustbox}{width=0.95\textwidth}
\begin{minipage}{\textwidth}
\centering\fontsize{8.5}{8.5}\selectfont
\setlength{\tabcolsep}{3.5pt}
\caption{Task-level statistics of $\mathrm{Robo_T}$, comprising 50 tasks and 2{,}166 simulated episodes used to construct action-mismatch pairs in AVM-Eval.}
\label{tab:tasks_ood_robot}
\begin{tabularx}{\textwidth}{@{}Yr@{\hspace{0.75em}}|@{\hspace{0.75em}}Yr@{}}\toprule
Task & Episodes & Task & Episodes \\
\midrule
adjust\_bottle & 33 & place\_can\_basket & 40 \\
beat\_block\_hammer & 37 & place\_cans\_plasticbox & 42 \\
blocks\_ranking\_rgb & 57 & place\_container\_plate & 43 \\
blocks\_ranking\_size & 49 & place\_dual\_shoes & 41 \\
click\_alarmclock & 11 & place\_empty\_cup & 50 \\
click\_bell & 11 & place\_fan & 53 \\
dump\_bin\_bigbin & 46 & place\_mouse\_pad & 44 \\
grab\_roller & 31 & place\_object\_basket & 44 \\
handover\_block & 32 & place\_object\_scale & 42 \\
handover\_mic & 52 & place\_object\_stand & 51 \\
hanging\_mug & 45 & place\_phone\_stand & 43 \\
lift\_pot & 45 & place\_shoe & 53 \\
move\_can\_pot & 55 & press\_stapler & 57 \\
move\_pillbottle\_pad & 57 & put\_bottles\_dustbin & 36 \\
move\_playingcard\_away & 41 & put\_object\_cabinet & 38 \\
move\_stapler\_pad & 41 & rotate\_qrcode & 50 \\
open\_laptop & 52 & scan\_object & 44 \\
open\_microwave & 56 & shake\_bottle & 33 \\
pick\_diverse\_bottles & 42 & shake\_bottle\_horizontally & 33 \\
pick\_dual\_bottles & 40 & stack\_blocks\_three & 43 \\
place\_a2b\_left & 44 & stack\_blocks\_two & 59 \\
place\_a2b\_right & 48 & stack\_bowls\_three & 43 \\
place\_bread\_basket & 47 & stack\_bowls\_two & 47 \\
place\_bread\_skillet & 41 & stamp\_seal & 54 \\
place\_burger\_fries & 47 & turn\_switch & 23 \\
\midrule
\multicolumn{3}{@{}l}{\textbf{Total: 50 tasks}} & \textbf{2{,}166} \\
\bottomrule
\end{tabularx}
\end{minipage}
\end{adjustbox}
\end{table}

\begin{table}[H]\centering
\begin{adjustbox}{width=0.90\textwidth}
\begin{minipage}{\textwidth}
\centering\fontsize{8.5}{8.5}\selectfont
\setlength{\tabcolsep}{3.5pt}
\caption{Task-level episode counts for $\mathrm{Gen}$: 19 tasks and 2{,}254 simulated episodes.}
\label{tab:tasks_ood_gen}
\begin{tabularx}{\textwidth}{@{}Yr@{\hspace{0.75em}}|@{\hspace{0.75em}}Yr@{}}\toprule
Task & Episodes & Task & Episodes \\
\midrule
Pick block by number & 170 & Pick block by color (camnoise) & 78 \\
Pick block by shape & 122 & Pick block by color (camocgen) & 114 \\
Pick object by type & 136 & Pick block by color (camposgen) & 128 \\
Pick specific object & 112 & Pick block by color (ctrlgen) & 112 \\
Pick billiard ball by color & 66 & Pick block by color (eefgen) & 98 \\
Pick block by color & 122 & Pick block by color (jointgen) & 188 \\
Pick block by color (instructgen) & 104 & Pick block by color (lightgen) & 98 \\
Pick block by size & 88 & Pick block by color (matgen) & 128 \\
Pick block by color (basegen) & 138 & Pick block by color (ref) & 120 \\
Pick block by color (camdropgen) & 132 &  &  \\
\midrule
\multicolumn{3}{@{}l}{\textbf{Total: 19 tasks}} & \textbf{2{,}254} \\
\bottomrule
\end{tabularx}
\end{minipage}
\end{adjustbox}
\end{table}

\begin{table}[H]\centering
\begin{adjustbox}{width=0.90\textwidth}
\begin{minipage}{\textwidth}
\centering\small
\setlength{\tabcolsep}{3.5pt}
\caption{Counts of action-mismatch pairs by numerical perturbation type (Appendix~\ref{app:local-numerical}). Each pair is counted once per type, so totals across types may exceed the number of distinct pairs.}
\label{tab:pert_numeric}
\begin{adjustbox}{max width=\textwidth}
\begin{tabular}{@{}ll cccccc@{}}\toprule
\multirow{2}{*}{Group} & \multirow{2}{*}{Operator} & \multicolumn{5}{c}{AVM-Eval} & AVM-Pref \\
\cmidrule(lr){3-7}\cmidrule(l){8-8}
 & & $\mathrm{Agi_A^{test}}$ & $\mathrm{Agi_B}$ & $\mathrm{Robo_T}$ & $\mathrm{Robo_D}$ & $\mathrm{Gen}$ & $\mathrm{Agi_A^{train}}$ \\
\midrule
\multirow{3}{*}{Translation perturbations} & Magnitude & 930 & 2{,}241 & 718 & 472 & 266 & 12{,}020 \\
 & Direction & 910 & 2{,}315 & 818 & 494 & 242 & 12{,}021 \\
 & Magnitude \& Direction & 963 & 2{,}277 & 742 & 473 & 262 & 12{,}283 \\
\cmidrule(l){1-8}
\multirow{3}{*}{Rotation perturbations} & Orientation Direction & 938 & 2{,}371 & 823 & 452 & 182 & 12{,}014 \\
 & Orientation Magnitude & 955 & 2{,}158 & 774 & 433 & 134 & 12{,}033 \\
 & Orientation Magnitude \& Direction & 943 & 2{,}167 & 828 & 457 & 160 & 12{,}142 \\
\cmidrule(l){1-8}
\multirow{3}{*}{\shortstack[l]{Gripper, Gaussian,\\and temporal order perturbations}} & Gripper Perturbation & 915 & 1{,}905 & 1{,}919 & 992 & 556 & 11{,}874 \\
 & Gaussian Perturbation & 2{,}549 & 6{,}648 & 1{,}155 & 612 & 504 & 33{,}274 \\
 & Temporal Order Permutation & 2{,}385 & 6{,}289 & 2{,}445 & 1{,}330 & 378 & 31{,}628 \\
\midrule
\multicolumn{2}{@{}l}{\textbf{Distinct pairs}} & \textbf{4{,}905} & \textbf{12{,}504} & \textbf{4{,}429} & \textbf{2{,}616} & \textbf{1{,}220} & \textbf{63{,}522} \\
\bottomrule\end{tabular}\end{adjustbox}\end{minipage}
\end{adjustbox}
\end{table}

\begin{table}[H]\centering
\begin{adjustbox}{width=0.90\textwidth}
\begin{minipage}{\textwidth}
\centering\small
\setlength{\tabcolsep}{3.5pt}
\caption{Counts of action-mismatch pairs by semantic perturbation group (Appendix~\ref{app:local-semantic}). Each pair is counted once per group; ``--'' denotes no available pairs.}
\label{tab:pert_semantic}
\begin{tabular*}{\textwidth}{@{\extracolsep{\fill}}l cccccc@{}}\toprule
\multirow{2}{*}{Group} & \multicolumn{5}{c}{AVM-Eval} & AVM-Pref \\
\cmidrule(lr){2-6}\cmidrule(l){7-7}
 & $\mathrm{Agi_A^{test}}$ & $\mathrm{Agi_B}$ & $\mathrm{Robo_T}$ & $\mathrm{Robo_D}$ & $\mathrm{Gen}$ & $\mathrm{Agi_A^{train}}$ \\
\midrule
Position offset & 3{,}655 & 3{,}429 & 2{,}540 & 2{,}119 & 1{,}084 & 46{,}345 \\
\cmidrule(l){1-7}
Gripper timing & 1{,}728 & 1{,}738 & 2{,}810 & 1{,}503 & -- & 23{,}272 \\
\cmidrule(l){1-7}
\begin{tabular}[c]{@{}l@{}}Mid-motion\\drift or freeze\end{tabular} & 1{,}806 & 1{,}810 & 1{,}641 & 1{,}245 & 1{,}094 & 23{,}194 \\
\midrule
\textbf{Distinct pairs} & \textbf{7{,}189} & \textbf{6{,}977} & \textbf{6{,}991} & \textbf{4{,}867} & \textbf{2{,}178} & \textbf{92{,}811} \\
\bottomrule\end{tabular*}\end{minipage}
\end{adjustbox}
\end{table}

\begin{figure}[H]
\centering
\includegraphics[width=0.92\textwidth]{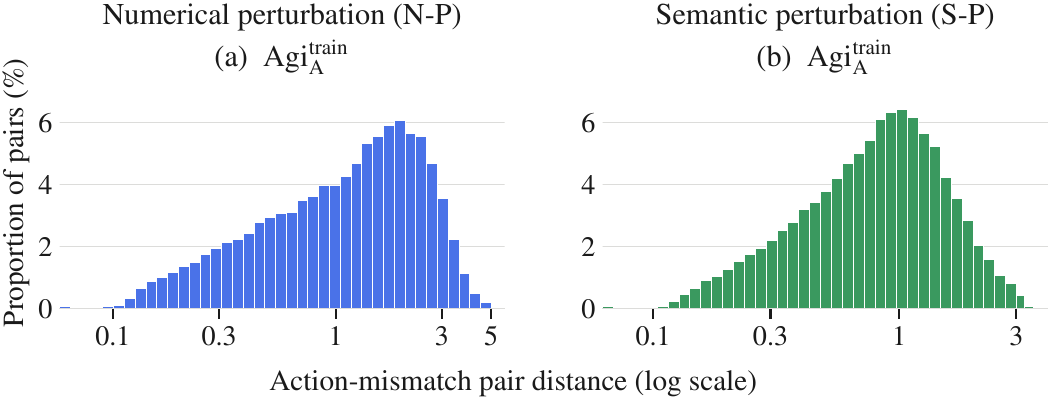}
\caption{Action-mismatch pair distances in AVM-Pref on $\mathrm{Agi_A^{train}}$, computed using Eq.~\ref{eq:pair-action-distance}. Panels (a) and (b) show numerical and semantic perturbations, respectively. The horizontal axes use a logarithmic scale.}
\label{fig:pref-action-pair-distance}
\end{figure}

\paragraph{Pairwise Separation Distributions.}
\label{app:pair-distances}
We characterize the separation between the positive and negative candidates using action differences for action-mismatch pairs and score differences for video-mismatch pairs.
The three distance definitions below are used to summarize the pair distributions.

\begin{figure}[t]
\centering
\includegraphics[width=0.92\textwidth]{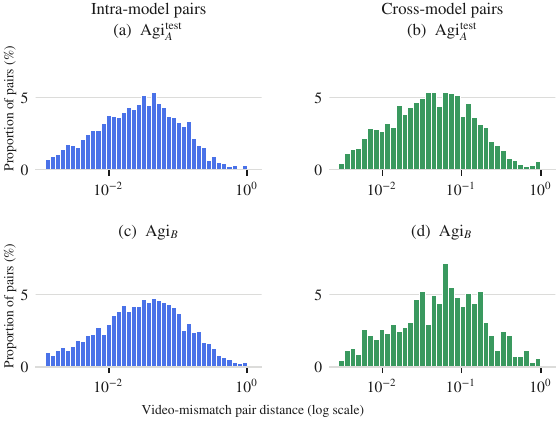}
\caption{Video-mismatch nDTW gaps in AVM-Eval (Eq.~\ref{eq:pair-eval-video-distance}). Rows: $\mathrm{Agi_A^{test}}$ and $\mathrm{Agi_B}$; columns: intra-model and cross-model pairs. Horizontal axes are logarithmic.}
\label{fig:eval-video-pair-distance}
\vspace{-6pt}
\end{figure}

\begin{figure}[t]
\centering
\includegraphics[width=0.87\textwidth]{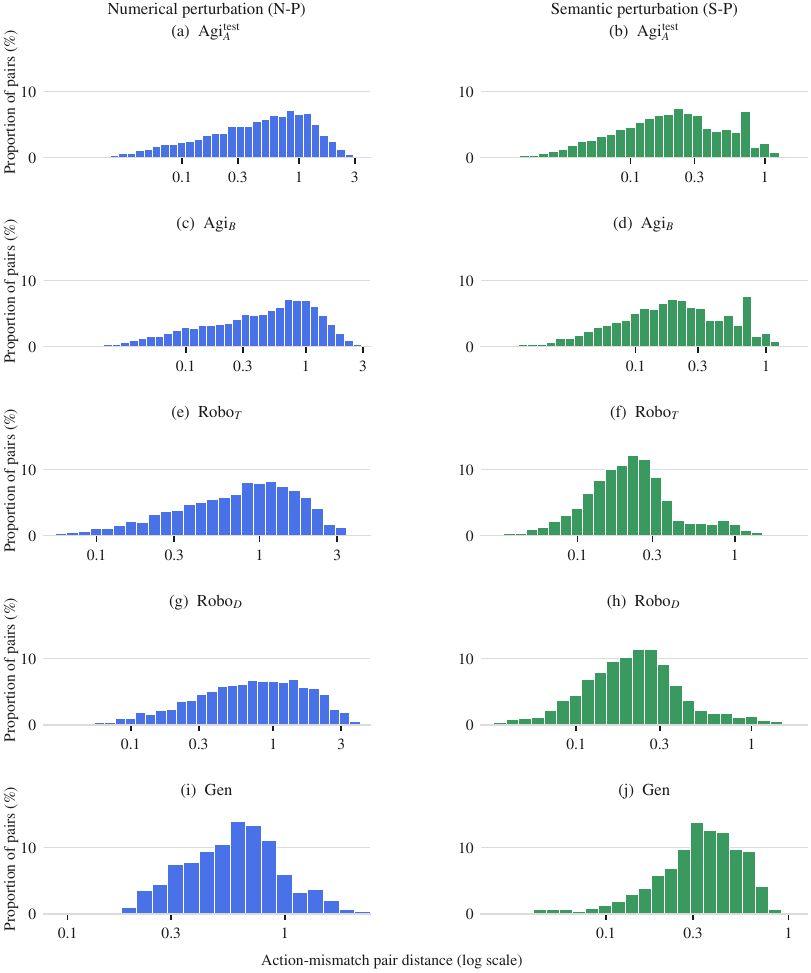}
\caption{Action-mismatch pair distances in AVM-Eval (Eq.~\ref{eq:pair-action-distance}). Rows: five source datasets; columns: numerical and semantic perturbations. Horizontal axes are logarithmic.}
\label{fig:eval-action-pair-distance}
\vspace{-6pt}
\end{figure}

\begin{figure}[t]
\centering
\includegraphics[width=0.87\textwidth]{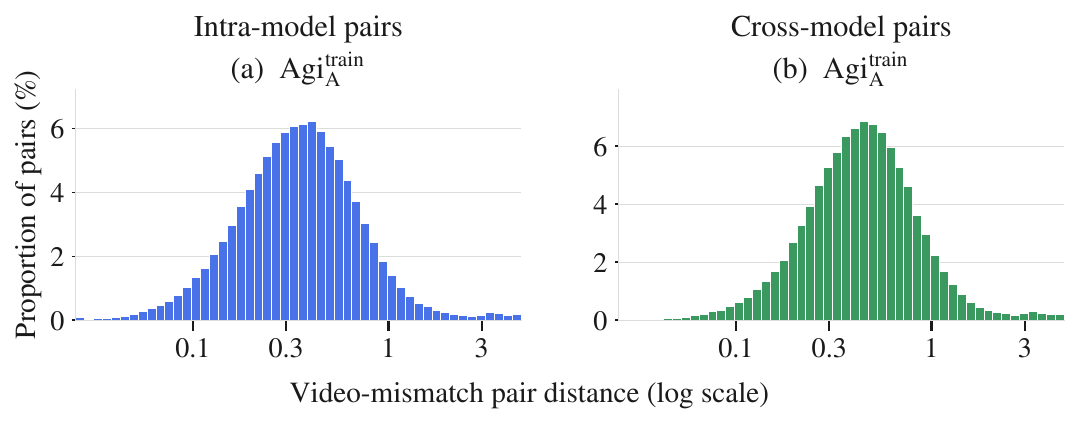}
\caption{Video-mismatch pair distances in AVM-Pref on $\mathrm{Agi_A^{train}}$ (Eq.~\ref{eq:pair-pref-video-distance}). Panels (a) and (b): intra-model and cross-model pairs. Horizontal axes are logarithmic.}
\label{fig:pref-video-pair-distance}
\vspace{-6pt}
\end{figure}

\begin{enumerate}[label=(\arabic*),leftmargin=1.9em,itemsep=4pt,topsep=4pt]

\item \textbf{Action-mismatch pairs in AVM-Eval and AVM-Pref.}
Let $\mathbf{a}^{+},\mathbf{a}^{-}\in\mathbb{R}^{K\times14}$ denote the two action chunks, where $K=12$ is the number of action transitions in a 13-frame clip.
Following Appendix~\ref{app:action_perturbation}, $t=1,\ldots,K$ indexes the transitions and $j=0,\ldots,13$ indexes the action dimensions.
We divide the difference in each dimension by its fixed standard deviation $\sigma_j$, estimated from the corresponding source dataset, and compute the root mean square over the entire chunk:
\begin{equation}
    d_{\mathrm{act}}(\mathbf{a}^{+},\mathbf{a}^{-})
    =\sqrt{\frac{1}{14K}\sum_{t=1}^{K}\sum_{j=0}^{13}
    \left(\frac{a^{+}_{t,j}-a^{-}_{t,j}}{\sigma_j}\right)^2}.
    \label{eq:pair-action-distance}
\end{equation}
This descriptive distance averages over all entries, whereas the filtering distance in Eq.~\ref{eq:local-distance} averages only over modified entries.
Figures~\ref{fig:eval-action-pair-distance} and~\ref{fig:pref-action-pair-distance} show the distributions for numerical and semantic perturbations in AVM-Eval and AVM-Pref, respectively.

\item \textbf{Video-mismatch pairs in AVM-Eval.}
Let $\hat{\mathbf{I}}^{+}$ and $\hat{\mathbf{I}}^{-}$ denote the positive and negative generated videos, and let $\mathbf{I}$ denote their shared ground-truth video.
We measure separation by the score difference:
\begin{equation}
    d_{\mathrm{vid}}^{\mathrm{Eval}}(\hat{\mathbf{I}}^{+},\hat{\mathbf{I}}^{-})
    =\left|\mathrm{nDTW}(\hat{\mathbf{I}}^{+},\mathbf{I})
    -\mathrm{nDTW}(\hat{\mathbf{I}}^{-},\mathbf{I})\right|.
    \label{eq:pair-eval-video-distance}
\end{equation}
Figure~\ref{fig:eval-video-pair-distance} reports this gap for intra-model and cross-model pairs on $\mathrm{Agi_A^{test}}$ and $\mathrm{Agi_B}$.

\item \textbf{Video-mismatch pairs in AVM-Pref.}
For each candidate video $\hat{\mathbf{I}}$, we collect the four quality metrics from Appendix~\ref{app:video-mismatch-pairs-for-AVM-Pre}, evaluated relative to the shared ground-truth video $\mathbf{I}$:
\[
    \mathbf{m}(\hat{\mathbf{I}},\mathbf{I})
    =\bigl[\mathrm{PSNR}^{\mathrm{glob}},\,
    \mathrm{PSNR}^{\mathrm{loc}},\,
    \mathrm{LPIPS}^{\mathrm{glob}},\,
    \mathrm{LPIPS}^{\mathrm{loc}}\bigr].
\]
For each metric $\ell\in\{1,2,3,4\}$, we estimate its standard deviation $\sigma_{\ell}^{\mathrm{VQ}}$ over all unique video observations appearing on either side of the preference pairs.
The same four standard deviations are used for intra-model and cross-model pairs.
We standardize each metric difference and take their root mean square:
\begin{equation}
    \begin{aligned}
    &\Delta_{\ell}^{\mathrm{VQ}}
    =\frac{m_{\ell}(\hat{\mathbf{I}}^{+},\mathbf{I})
    -m_{\ell}(\hat{\mathbf{I}}^{-},\mathbf{I})}{\sigma_{\ell}^{\mathrm{VQ}}},\\
    &d_{\mathrm{vid}}^{\mathrm{Pref}}(\hat{\mathbf{I}}^{+},\hat{\mathbf{I}}^{-})
    =\sqrt{\frac{1}{4}\sum_{\ell=1}^{4}
    \left(\Delta_{\ell}^{\mathrm{VQ}}\right)^2}.
    \end{aligned}
    \label{eq:pair-pref-video-distance}
\end{equation}
Figure~\ref{fig:pref-video-pair-distance} shows the distributions of this distance for intra-model and cross-model pairs on $\mathrm{Agi_A^{train}}$.
\end{enumerate}


\section{JEPA-IDM Implementation Details}
\label{app:jepa-idm}

JEPA-IDM shares ACWM-RM's video preprocessing, camera-frame action parameterization, frozen V-JEPA 2.1 encoder, and predictor backbone (Appendix~\ref{app:predictor}).
It predicts actions instead of visual features and scores candidates using the IDM baselines' action-distance formulation instead of a learned scalar score head.

\paragraph{Predictor Inputs.}
\label{sec:idm-input}
For a clip $\mathbf I=(I_0,\ldots,I_K)$, action $a_t$ corresponds to the transition from $I_{t-1}$ to $I_t$.
Following Appendix~\ref{app:predictor}, we independently encode each frame and project its visual features:
\begin{equation}
    z_t=\operatorname{LN}(E(I_t))\in\mathbb R^{N\times D},
    \qquad U_t^z=f_z(z_t)\in\mathbb R^{N\times d_P}.
    \label{eq:jepa-idm-visual-input}
\end{equation}
We use two learnable tokens $e_L,e_R\in\mathbb R^{d_P}$ as action readout tokens for the left and right arms, respectively.
The initial block contains only the visual tokens of $I_0$.
For each subsequent frame $I_t$, we append the two readout tokens after its visual tokens:
\begin{equation}
\begin{aligned}
    \mathbf X_0^{\mathrm{IDM}}&=U_0^z,\\
    \mathbf X_t^{\mathrm{IDM}}
    &=\big[U_t^z,e_L,e_R\big],\qquad t=1,\ldots,K,\\
    \mathbf X^{\mathrm{IDM}}
    &=\big[\mathbf X_0^{\mathrm{IDM}},\ldots,\mathbf X_K^{\mathrm{IDM}}\big].
\end{aligned}
    \label{eq:jepa-idm-input-blocks}
\end{equation}
For a two-frame input, the sequence is therefore $[U_0^z,U_1^z,e_L,e_R]$, with both arm readout tokens placed after the two frames' visual tokens.
Block-causal attention lets block $t$ access blocks $0,\ldots,t$, including both transition frames $I_{t-1}$ and $I_t$ when $t\geq1$.
During preference training and evaluation, all visual blocks come from the candidate's own video; the candidate action chunk is used only to compute its score.

\paragraph{Predictor Outputs.}
\label{sec:idm-readout}
Let $h_{t,r}\in\mathbb R^{d_P}$ denote the predictor's output at arm $r$'s readout token appended to block $t$.
A shared MLP $g:\mathbb R^{d_P}\rightarrow\mathbb R^7$ directly maps each readout feature to an action:
\begin{equation}
    \tilde a_t^r=g(h_{t,r}),
    \qquad r\in\{L,R\},\quad t=1,\ldots,K,
    \label{eq:jepa-idm-action-readout}
\end{equation}
The seven components are a 3D translation increment, a 3D rotation-vector increment, and a gripper-opening increment.
Translation and rotation increments are expressed in the head-camera coordinate frame, while gripper increments retain their original units.
Per-dimension standardization is applied to their residuals in Equations~\ref{eq:jepa-idm-pretraining-loss} and~\ref{eq:jepa-idm-preference-score}.
The initial block has no action readout; each subsequent block produces one action per arm for its incoming transition.
Thus, the output for $B$ clips containing $K+1$ frames has shape $[B,K,2,7]$; each 13-frame clip produces 12 bimanual actions.

\paragraph{Action-Prediction Pretraining.}
We pretrain the predictor on real video-action pairs from $\mathrm{Agi_A^{train}}$ by minimizing the standardized Smooth-L1 loss between the predicted actions and their ground-truth counterparts:
\begin{equation}
    \mathcal{L}_{\mathrm{IDM}}
    =\frac{1}{14K}\sum_{t=1}^{K}\sum_{r\in\{L,R\}}\sum_{j=1}^{7}
    \ell_{\mathrm{SL1}}\!\left(
    \frac{\tilde a_{t,j}^{r}-a_{t,j}^{r}}{\sigma_j^{r}+10^{-8}}
    \right).
    \label{eq:jepa-idm-pretraining-loss}
\end{equation}
Here, $\sigma_j^r$ is the training-set standard deviation of the corresponding action component; the same fixed statistics are used in preference training.
The visual encoder remains frozen.
We use the same training hyperparameters as ACWM-RM's dynamics pretraining (Appendix~\ref{app:predictor}).

\paragraph{Preference Training.}
We initialize from action-prediction pretraining and fine-tune the predictor on AVM-Pref, keeping the visual encoder frozen.
For the preferred ($+$) and dispreferred ($-$) candidates, the reward model infers actions from each candidate's video and computes its score using the associated actions:
\begin{equation}
    R^{\pm}
    =-\frac{1}{14K}\sum_{t=1}^{K}\sum_{r\in\{L,R\}}\sum_{j=1}^{7}
    \ell_{\mathrm{SL1}}\!\left(
    \frac{\tilde a_{t,j}^{r,\pm}-a_{t,j}^{r,\pm}}{\sigma_j^{r}+10^{-8}}
    \right).
    \label{eq:jepa-idm-preference-score}
\end{equation}
We optimize the Bradley--Terry loss over preference pairs from AVM-Pref:
\begin{equation}
    \mathcal{L}_{\mathrm{BT}}
    =-\mathbb{E}_{(q^+,q^-)\sim\mathrm{AVM\text{-}Pref}}
    \left[\log\sigma\!\left(R^{+}-R^{-}\right)\right],
    \label{eq:jepa-idm-bt-loss}
\end{equation}
where $q^{\pm}$ denote the preferred and dispreferred candidates and $\sigma$ is the sigmoid function.
We use the same optimizer settings, batch size, training duration, and schedule as ACWM-RM's preference training (Appendix~\ref{app:preference-training}).
All trainable JEPA-IDM parameters use $0.05$ times ACWM-RM's score-head learning rate throughout preference training, including warmup, with a peak learning rate of $2\times10^{-5}$.

\end{document}